\documentclass[10pt,twocolumn,letterpaper]{article}

\usepackage{cvpr}              

\newcommand{\myparagraph}[1]{\vspace{4pt}\noindent{\bf #1}}

\definecolor{cvprblue}{rgb}{0.21,0.49,0.74}
\usepackage[pagebackref,breaklinks,colorlinks,allcolors=cvprblue]{hyperref}

\def\paperID{****} 
\def\confName{CVPR}
\def\confYear{2026}

\usepackage[table]{xcolor}   
\definecolor{bestcolor}{HTML}{FFC5C5}   
\definecolor{secondcolor}{HTML}{FFEBd8} 

\newcommand{\best}[1]{%
  \multicolumn{1}{>{\columncolor{bestcolor}}c}{\bfseries #1}%
}

\newcommand{\secbest}[1]{%
  \multicolumn{1}{>{\columncolor{secondcolor}}c}{#1}%
}

\title{Decoupled Generative Modeling for Human-Object Interaction Synthesis}

\author{
Hwanhee Jung\textsuperscript{1}\hspace{0.4em}
Seunggwan Lee\textsuperscript{1}\hspace{0.3em}
Jeongyoon Yoon\textsuperscript{1}\hspace{0.4em}
SeungHyeon Kim\textsuperscript{1}\hspace{0.4em} \\
Giljoo Nam\textsuperscript{2}\hspace{0.4em}
Qixing Huang\textsuperscript{3}\hspace{0.4em}
Sangpil Kim\textsuperscript{1}\footnotemark[1] \\
\vspace{-.6em}
\\
\vspace{-1.em}
\textsuperscript{1}Korea University \quad
\textsuperscript{2}Meta \quad
\textsuperscript{3}The University of Texas at Austin \quad
}

\begin{document}
\maketitle
\def\thefootnote{*}\footnotetext{Corresponding author.}
\begin{abstract}
Synthesizing realistic human-object interaction (HOI) is essential for 3D computer vision and robotics, underpinning animation and embodied control. Existing approaches often require manually specified intermediate waypoints and place all optimization objectives on a single network, which increases complexity, reduces flexibility, and leads to errors such as unsynchronized human and object motion or penetration. To address these issues, we propose Decoupled Generative Modeling for Human-Object Interaction Synthesis (\textbf{DecHOI}), which separates path planning and action synthesis. A trajectory generator first produces human and object trajectories without prescribed waypoints, and an action generator conditions on these paths to synthesize detailed motions. 
To further improve contact realism, we employ adversarial training with a discriminator that focuses on the dynamics of distal joints.
The framework also models a moving counterpart and supports responsive, long‑sequence planning in dynamic scenes, while preserving plan consistency. 
Across two benchmarks, FullBodyManipulation and 3D-FUTURE, DecHOI surpasses prior methods on most quantitative metrics and qualitative evaluations, and perceptual studies likewise prefer our results.
\end{abstract}    
\vspace{-2em}
\section{Introduction}
\label{sec:intro}
Realistic human-object interaction synthesis (HOI) is a fundamental task with broad impact in computer vision and robotics~\cite{bhatnagar2022behave, xu2025interact, liu2022hoi4d, cao2023detecting, chen2023detecting}. These capabilities form the basis of modern 3D systems, enabling human motion animation and humanoid control~\cite{zhu2023human, xue2025human, liu2025core4d}. Synthesizing interaction requires perceiving the pose of the object and awareness of the target goal, followed by the generation of a plausible sequence of human joint configurations that performs the required manipulation safely and intentionally~\cite{antoun2023human, yang2024lemon, yang2024f, xu2025intermimic}. Recent approaches~\cite{li2024controllable, wu2025human} rely on interpretable natural language instructions to specify tasks, yet producing interactions that are faithful to the prompt while keeping realism and diversity remains a difficult challenge.
\begin{figure}[t]
    \centering
    \includegraphics[width=\linewidth]{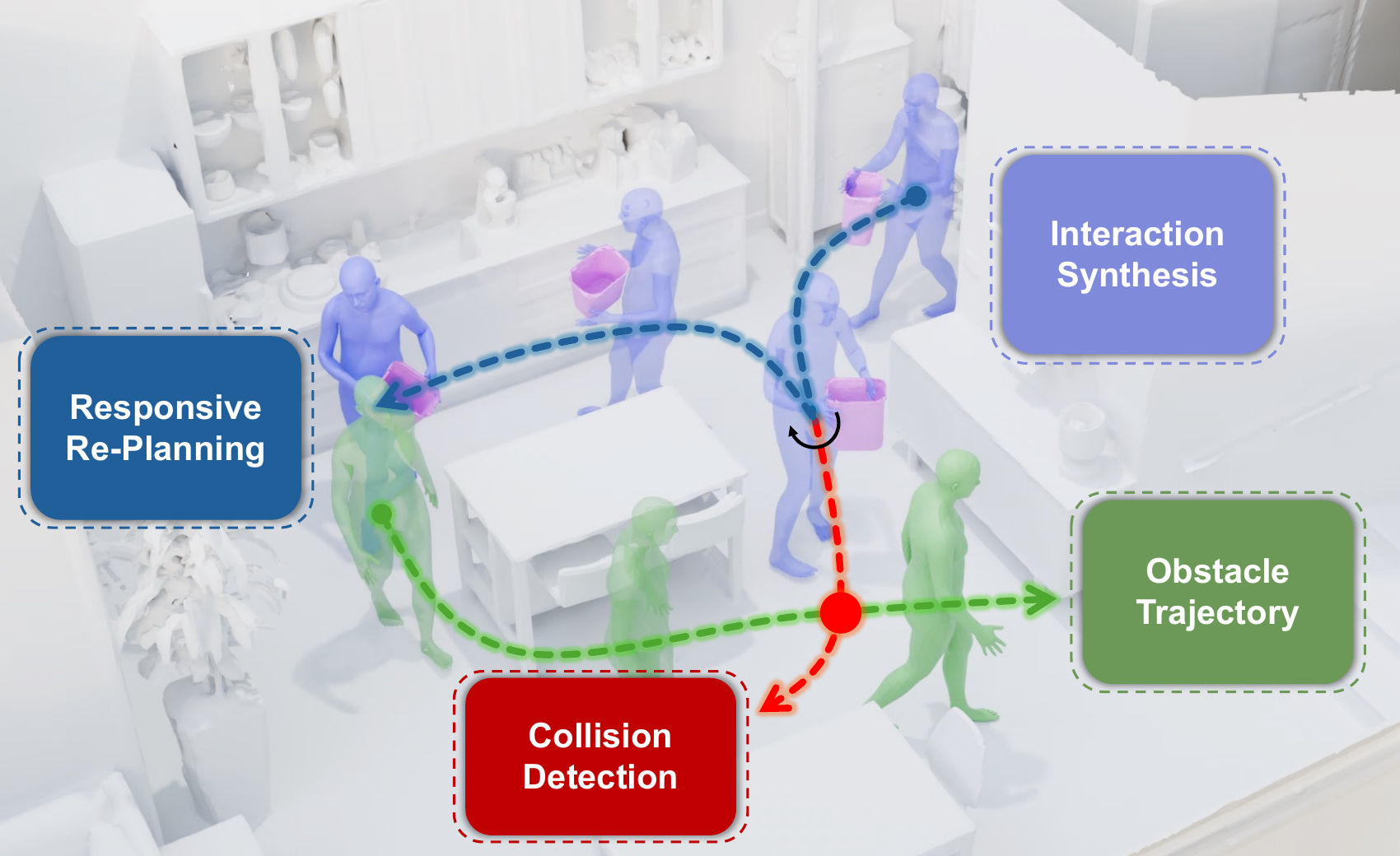}
    \vspace{-1.8em}
    \caption{
    Overview of DecHOI for dynamic human-object interaction synthesis. The framework decouples trajectory planning and interaction synthesis, enabling collision detection and responsive re-planning for realistic, contact-consistent motion. 
    }
    \label{fig:teaser}
    \vspace{-2em}
\end{figure}
Prior methods~\cite{li2024controllable, wu2025human} generate 4D motion sequences of humans and objects by conditioning a diffusion model on path waypoints and text instructions. While these methods provide a strong foundation for interaction modeling that is capable of producing desired behaviors, this design introduces two challenges: i) reliance on specified intermediate waypoints that the human must follow, and ii) high optimization complexity from assigning the entire objective to a single denoising network. Specifically, for the first challenge, at inference time, the user must manually provide not only the start and goal points but also intermediate waypoints. Such reliance on externally supplied constraints narrows the model's generative scope and reduces autonomy, while introducing procedural burden for the user. For the second challenge, HOI requires the model to simultaneously resolve the motion of both the human and the object in every frame. Assigning all objectives to a single network elevates optimization complexity and risks practical intractability. The resulting complexity often produces unsynchronized interactions, such as hovering or penetration of objects, which harms plausibility and degrades overall quality. These observations motivate us to design a new intermediate representation that is simple, flexible, informative, and still enables interactions that remain faithful to the instruction and plausible for HOI generation.

To this end, we propose Decoupled Generative Modeling for Human-Object Interaction Synthesis (DecHOI), a novel framework that generates trajectories and motions as separate processes. In DecHOI, a trajectory generator (TG) first produces human and object trajectories, and an action generator (AG) then synthesizes detailed actions conditioned on the generated paths. This decoupling allows the TG to produce diverse and accurate paths without reliance on specified waypoints, while the AG focuses exclusively on human actions and object motion, capturing more fine-grained details. Partitioning the model into two lightweight expert networks reduces the optimization complexity for each branch and mitigates unsynchronized interactions.

Another key factor for human-level interaction is precise hand and foot control~\cite{hao2024hand, mao2022contact}. To improve the robustness of hand and foot contact and reduce undesired penetration, we introduce adversarial training applied to the object and to the distal joints. A compact discriminator focuses on hand-object interaction signals and foot dynamics, shaping the learning distribution and enabling realistic control.

For applications, the previous model~\cite{li2024controllable} enables 3D scene-aware long-sequence interaction synthesis but remains restricted to static environments. To overcome this limitation, our framework models a moving counterpart and equips the human agent to either avoid the counterpart or wait, allowing it to respond to dynamic environments. 
This improves practicality in dynamic scenes by aligning actions with the plan and adapting to moving obstacles.

To validate the effectiveness of DecHOI, we evaluated our method on the \textit{FullBodyManipulation}~\cite{li2023object} and observed superior performance on most quantitative metrics and qualitative assessments. We also report results on unseen objects from the \textit{3D-FUTURE}~\cite{fu20213d}, demonstrating robustness and generalizability. For long-sequence interaction synthesis, we present scenarios in indoor environments furnished with diverse objects, where the human agent interacts with a moving counterpart and reaches various goal points by either avoiding the counterpart or waiting. The summary of our contributions is as follows.

\begin{itemize} 
\item We propose DecHOI, a decoupled framework that separates trajectory generation from fine‑grained action synthesis, reducing optimization complexity and removing the need for manual intermediate waypoints.
\item We improve coordination between distal joints and an object through adversarial training with a compact discriminator, thereby reducing interpenetration.
\item We enable responsive planning with a long-sequence planner that dynamically adapts to moving counterparts, supporting scene-aware interaction.
\item Experiments conducted on various benchmarks demonstrate that DecHOI achieves state-of-the-art performance, surpassing previous methods in terms of realism, accuracy, and diversity.
\end{itemize}

\section{Related Work}
\label{sec:related}

\subsection{Human Action Generation}
Human action generation aims to synthesize realistic 3D motion sequences from conditioning signals such as text descriptions, action labels, and joint-level constraints~\cite{meng2025rethinking, zhang2025energymogen, fan2024freemotion, zou2024parco, jiang2023motiongpt}. T2M~\cite{guo2022generating} samples a motion length from text and then synthesizes the motion with a temporal VAE over motion snippet codes. MDM~\cite{tevet2023human} adapts classifier-free diffusion~\cite{ho2022classifier} with a transformer backbone for high-fidelity motion, and PriorMDM~\cite{shafir2024human} treats a pre-trained diffusion model as a generative prior to enable controllable composition. ACMDM~\cite{meng2025absolute} uses absolute global joint coordinates in a streamlined transformer diffusion model, improving fidelity and text alignment. OmniControl~\cite{xie2024omnicontrol} adds spatial constraints for any joint at any time using analytic guidance with a learned realism prior, while MoMask~\cite{guo2024momask} enables text-guided temporal inpainting via masked generation with hierarchical motion tokens. However, these works primarily address single human motion and neither manipulate objects nor perceive or adapt to the surrounding scene~\cite{wang2025hsi, wang2024move, li2024genzi}. In contrast, we target scene-aware human-object interaction that avoids obstacles and manipulates objects in accordance with the specified intent.

\subsection{Human-Object Interaction Synthesis}
Human-object interaction synthesis (HOI) takes text instructions or object state as input and generates purposeful 3D human-object motion~\cite{xue2025guiding, jia2025primhoi, zhang2025interactanything, song2024hoianimator}. 
For example, OMOMO~\cite{li2023object} conditions on full per-frame object motion and enforces contact by integrating hand kinematics during whole body generation.
CHOIS~\cite{li2024controllable} and HOIFHLI~\cite{wu2025human} show that per-frame object conditioning is unnecessary because language instruction paired with sparse 3D object waypoints, together with the initial human and object states, stabilizes intended behaviors.
As a complementary effort, CG-HOI~\cite{diller2024cg} jointly generates human and object motion with contact cues, using contact-guided conditioning to improve realism. Despite these gains, a single network that models trajectory, pose, and contact is still hard to optimize and often produces desynchronization and contact artifacts. We instead decouple trajectory and action generation and apply contact-aware adversarial training, yielding robust synchronized HOI.
\begin{figure*}[t]
    \centering
    \includegraphics[width=1.0\textwidth]{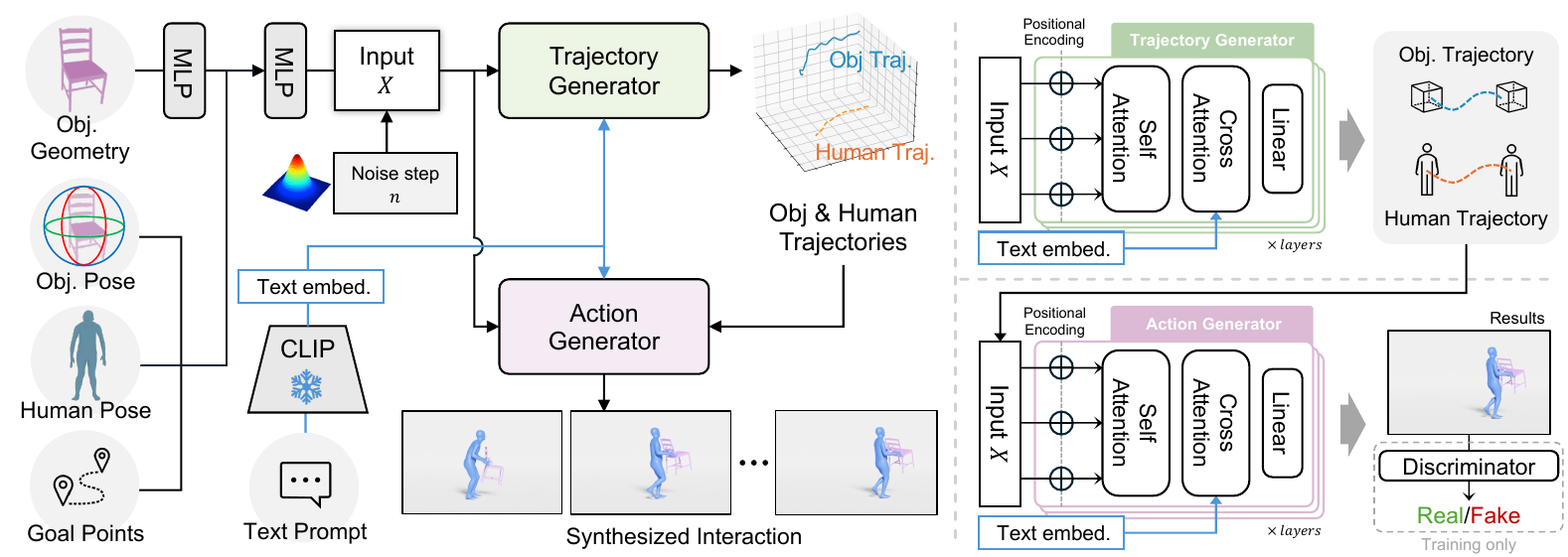}
    \vspace{-2.em}
    \caption{
    Architecture of DecHOI showing the decoupled trajectory and action generation process. 
    Conditioned on the text instruction, geometry, current human and object poses, and a goal point, the trajectory generator plans paths, while the action generator produces joint motions on these paths to yield synchronized, contact-aware interactions. The right panels detail the Trajectory and Action Generators.
    }

    \label{fig:main_arch}
    \vspace{-1.em}
\end{figure*}


\subsection{Path-conditioned HOI}
Path-conditioned HOI first specifies a long-term trajectory via waypoints, object paths, or scene-level guidance, and then synthesizes it at the motion level~\cite{yi2024generating, cen2024generating, lee2023locomotion}. 
CHOIS generates synchronized long-horizon human-object motion in 3D scenes by conditioning diffusion on sparse object waypoints obtained from Habitat~\cite{savva2019habitat, szot2021habitat} for the given start and target positions. NIFTY~\cite{kulkarni2024nifty} couples an object-conditioned motion diffusion model with a learned interaction field that evaluates distances to a feasible interaction manifold and guides sampling toward contact-consistent behaviors.
OMOMO~\cite{li2023object} takes per-frame object motion data as input and synthesizes manipulation by denoising, incorporating hand motion to maintain contact during whole body generation.
However, in inference, these formulations retain a fixed plan or guidance and therefore do not adapt to external scene changes. By contrast, our method updates the plan, enabling adaptation to external scene changes while preserving high-level intent.

\section{Method}
\label{sec:method}
Given an instruction, current human and object poses, and a goal point, our framework synthesizes an interaction that moves the object to the goal as instructed. As illustrated in Fig.~\ref{fig:main_arch}, we adopt a decoupled generation architecture with a trajectory generator and an action generator (Sec.~\ref{sec:decoupled}) to reduce reliance on explicit waypoints and simplify the optimization objective. To improve contact realism and reduce interpenetration, we employ adversarial training with a compact discriminator that focuses on the dynamics of distal joints (Sec.~\ref{sec:adversarial}). In Sec.~\ref{sec:Training}, we present the overall optimization procedure and the inference-time guidance strategies for our method. 
\vspace{-0.2em}
\subsection{Background: Denoising Diffusion Model}
\label{sec:background}
\vspace{-0.2em}
Denoising Diffusion Probabilistic Models (DDPMs)~\cite{ho2020denoising, tevet2023human} are generative models that learn to approximate complex data distributions by gradually adding noise and then removing it. This paradigm has become dominant in recent work on human action generation~\cite{tevet2023human, shafir2024human, meng2025absolute, xie2024omnicontrol} and interaction synthesis~\cite{li2023object, li2024controllable, wu2025human, xu2023interdiff}. The forward process progressively corrupts a clean sample $\mathbf{x}_0$ by adding Gaussian noise over $N$ steps, defined as:
\vspace{-0.3em}
\begin{equation}
\label{eq:back_forward_proc}
q(\mathbf{x}_n \mid \mathbf{x}_{0}) = \mathcal{N}\!\bigl(\mathbf{x}_n; \sqrt{\bar{\alpha}_n}\mathbf{x}_{0}, (1-\bar{\alpha}_n) \mathbf{I}\bigr),
\end{equation}
where $\alpha_n = 1-\beta_n$ and $\bar{\alpha}_n = \prod_{s=1}^{n} \alpha_s$. The forward process is a Markov chain with Gaussian transitions. As $n$ increases, $\mathbf{x}_n$ converges to an isotropic Gaussian. The reverse process learns to invert this noising procedure with a parameterized model $p_\theta$ that progressively denoises $\mathbf{x}_n$ to recover clean data:
\vspace{-0.5em}
\begin{equation}
\label{eq:back_reverse_proc}
p_\theta(\mathbf{x}_{n-1}\mid \mathbf{x}_n, \mathbf{c}) = \mathcal{N}\!\bigl(\mathbf{x}_{n-1}; \boldsymbol{\mu}_\theta(\mathbf{x}_n, n, \mathbf{c}), \sigma_n^2\mathbf{I}\bigr),
\end{equation}
where $\mathbf{c}$ denotes conditioning inputs such as text embeddings, object geometry, or start and goal states. Following prior formulations~\cite{li2024controllable, wu2025human}, our denoising network predicts the clean sequence rather than the noise:
\vspace{-0.2em}
\begin{equation}
\label{eq:back_reverse_proc}
\mathcal{L} = \mathbb{E}_{\mathbf{x}_0, n \sim[1, N]} \| \hat{\mathbf{x}}_0 - \mathbf{x}_0 \|_1.
\end{equation}

Predicting $\hat{\mathbf{x}}_0$ has been found to yield better results for motion data and allows guidance losses to be applied at each denoising step.

\subsection{Decoupled Generative Modeling}
\label{sec:decoupled}
Jointly estimating and optimizing human and object poses is non-trivial for a single network to handle. We address this challenge with decoupled generative modeling that separates trajectory generation from action generation. At the start of the pipeline the inputs are the object pose sequence $P_o \in \mathbb{R}^{T \times 12}$ with $T$ frames, where each frame contains the global 3D position and a 9-parameter relative rotation matrix, and the human pose sequence $P_h \in \mathbb{R}^{T \times D_h}$, where $D_h$ comprises global joint coordinates and per-joint 6D rotation parameters. To ensure accurate interaction and instruction following, inputs also include object geometry $B \in \mathbb{R}^{1024\times 3}$ represented as a Basis Point Set (BPS)~\cite{prokudin2019efficient} and a text instruction.

\myparagraph{Trajectory Generator.}
We formulate the trajectory generator (TG) as a conditional denoising diffusion model~\cite{ho2020denoising}. Given an input sequence, we apply a forward noising process for $N$ steps following a Markov chain. For conditioning, the human and object poses of the start frame remain clean, and the object position in the end frame is also kept clean to set the goal point.
Then these noisy data are concatenated with the geometry embedding $F_{\text{obj}} \in \mathbb{R}^{D}$, computed by a simple MLP encoder from $B$.
We obtain the conditioning input $X \in \mathbb{R}^{T \times (12 + D_h + D)}$ for the denoising network. A transformer-based denoising network of TG generates both global object and human trajectories that are faithful to the instruction. For example, given the instruction \textit{“Lift the chair, move the chair, and put down the chair,”} the object trajectory rises in the early part of the sequence and descends as it approaches the goal point. 
To learn this alignment, we incorporate a text embedding $F_{\text{text}} \in \mathbb{R}^{D}$ obtained from a CLIP encoder~\cite{radford2021learning} into the model inputs. 
In contrast to prior methods~\cite{li2024controllable, wu2025human} that condition by concatenating the embedding along the sequence dimension, we use a cross-attention layer to apply the conditioning explicitly, which yields stronger alignment between text and motion features.
Afterward, the TG yields continuous 3D paths for the object $\hat{\mathcal{T}}_o \in \mathbb{R}^{T \times 3}$ and the human $\hat{\mathcal{T}}_h \in \mathbb{R}^{T \times 3}$.

\myparagraph{Action Generator.}
Once reliable human and object trajectories have been generated, we use them as richer conditions. The action generator (AG) receives the same input $X$ as the trajectory generator (TG), except that the noisy global positions of the object and the human root in all $T$ frames are replaced with the trajectories produced by TG. This dense conditioning provides stronger priors and reduces the complexity of the learning objective. AG is a lightweight diffusion model that employs a transformer-based denoising network with the same architecture as TG. Given the conditioned noisy input, it generates the full pose of the object $\hat{P}_o \in \mathbb{R}^{T \times 12}$ and the pose of the human joints $\hat{P}_h \in \mathbb{R}^{T \times D_h}$. The human pose generated is then used for parametric human modeling, and we apply SMPL-X~\cite{pavlakos2019expressive} to reconstruct the human mesh and pose.

\subsection{Adversarial Training for Distal Joints}
\label{sec:adversarial}
Realistic human-object interaction (HOI) requires not only plausible motion but also reliable contact~\cite{cseke2025pico, gu2024contactgen, jiang2024scaling}. Language instructions such as \textit{“Lift the chair”} or \textit{“Kick the box”} rely on distal joints in the hands and feet. 
Existing models often cause the hands and feet to drift into empty space or intersect objects~\cite{yuan2023physdiff, xu2025interact, mao2022contact}.
To address these issues, we introduce an adversarial training mechanism that regularizes contact by focusing on the distal joints.

\begin{figure}[t]
    \centering
    \includegraphics[width=\linewidth]{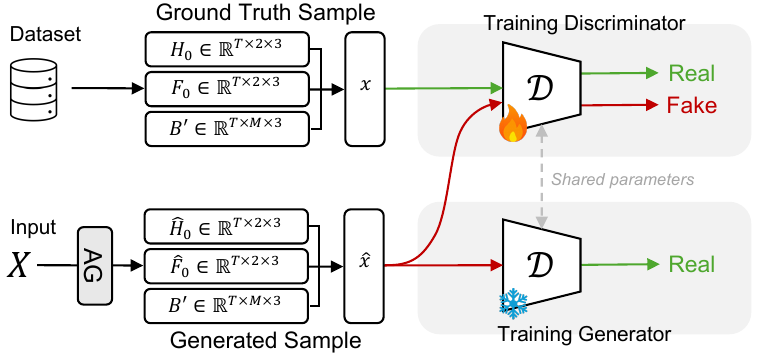}
    \vspace{-1.5em}
    \caption{
    Adversarial module of DecHOI, where a hand and foot-focused discriminator contrasts real and generated interactions to enhance contact realism.
    }
    \label{fig:discriminator}
    \vspace{-1.5em}
\end{figure}

In general, the most reliable cue to decide whether an input is real or fake is the distance between the object surface and the distal joints~\cite{mao2022contact, cseke2025pico, huang2022capturing, tripathi2023deco}. In ground truth, these distances are small due to complete contact, whereas generated results often exhibit larger gaps. Optimizing the generator so that the discriminator cannot distinguish real from fake therefore acts as a regularizer that drives contact toward completeness. Motivated by this observation, we design a compact discriminator $\mathcal{D}$ that differentiates real and fake using distal joint kinematics together with object geometry. As shown in Fig.~\ref{fig:discriminator}, $\mathcal{D}$ receives  global coordinates of the hands $H \in \mathbb{R}^{T \times 2 \times 3}$ and feet $F \in \mathbb{R}^{T\times 2 \times 3}$ over all $T$ frames, obtained by forward kinematics from the 6D joint rotations. In addition, we sample $M$ points from the geometry of the object $B$ to obtain $B’ \in \mathbb{R}^{T \times M \times 3}$ and apply the relative rotation and translation for each frame. For real data, we feed clean inputs $x$, and for fake data, we use the outputs $\hat{x}$ produced by AG. The discriminator is trained with the following loss:
\vspace{-1em}
\begin{equation}
\label{eq:d_hinge_loss}
\begin{split}
\mathcal{L}_{\mathcal{D}}
&= \frac{1}{T}\sum_{t=1}^{T}\big([1-s^{(r)}_t]_+ + [1+s^{(f)}_t]_+\big),\\
s^{(r)}_t &= \mathcal{D}(x)_t,\quad s^{(f)}_t = \mathcal{D}(\hat x)_t,
\end{split}
\end{equation}
where $s^{(r)}_t$ and $s^{(f)}_t$ denote the scores for real and fake data at frame $t$.
After completing the discriminator training stage, we train AG to fool the discriminator by minimizing the following generator objective:

\vspace{-0.4em}
\begin{equation}
\label{eq:g_hinge_loss}
\mathcal{L}_G = - \frac{1}{T} \sum_{t=1}^{T} s^{(f)}_t.
\end{equation}

Through adversarial training, the network enforces higher contact fidelity and synthesizes realistic interactions.

\begin{table*}[t]
\begin{center}
\resizebox{\textwidth}{!}{%
\begin{tabular}{@{}lcccccccccccccccc@{}} 
\toprule
& \multicolumn{2}{c}{Condition Matching} 
& \multicolumn{5}{c}{Human Motion Quality} 
& \multicolumn{5}{c}{Interaction Quality} 
& \multicolumn{4}{c}{GT Difference} \\ 
\cmidrule(lr){2-3} \cmidrule(lr){4-8} \cmidrule(lr){9-13} \cmidrule(lr){14-17}
Methods & $T_s$$\downarrow$ & $T_e$$\downarrow$ & $H_{\text{feet}}$$\downarrow$ & $FS$$\downarrow$ & $R_{prec}$$\uparrow$ & $FID$$\downarrow$ & $DIV$$\rightarrow$ & $C_{\text{prec}}$$\uparrow$ & $C_{\text{rec}}$$\uparrow$ & $C_{F1}$$\uparrow$ & $P_{\text{hand}}$$\downarrow$ & $P_{\text{body}}$$\downarrow$ & MPJPE$\downarrow$ & $T_{\text{root}}$$\downarrow$ & $T_{\text{obj}}$$\downarrow$ & $O_{\text{obj}}$$\downarrow$ \\ 
\midrule
Lin-OMOMO~\cite{li2023object} & - & - & 10.04 & 0.44 & 0.45 & 16.81 & 6.65 & 0.75 & 0.56 & 0.59 & 0.70 & 0.65 & \secbest{17.45}  & \secbest{29.20} & 75.65 & -\\
Pred-OMOMO~\cite{li2023object} & 2.34 & 9.66 & 6.99 & \best{0.38} & 0.63 & 13.01 & 6.68 & 0.67 & 0.48 & 0.52 & 0.59 & \secbest{0.60} & 23.63  & 49.12 & 51.86 & 1.23 \\
CHOIS~\cite{li2024controllable} & 1.92 & 8.01 & 6.15 & \secbest{0.41} & \secbest{0.68} & \secbest{1.58} & 8.31 & 0.74 & 0.53 & 0.58 & 0.66 & 0.61 & 18.86  & 44.04 & 51.86 & 1.23 \\
HOIFHLI~\cite{wu2025human} & \secbest{1.73} & \secbest{7.65} & \secbest{4.77} & \best{0.38} & 0.62 & 2.06 & \secbest{8.55} & \secbest{0.77} & \secbest{0.61} & \secbest{0.64} & \secbest{0.58} & 0.61 & 19.31 & 40.87 & \secbest{50.96} & \secbest{1.18} \\
\midrule
\textbf{Ours (DecHOI)} & \best{1.59} & \best{6.91} & \best{4.42} & \best{0.38} & \best{0.72} & \best{0.33} & \best{8.86} & \best{0.80} & \best{0.64} & \best{0.67} & \best{0.53} & \best{0.54} & \best{15.27} & \best{25.47} & \best{22.96} & \best{0.86}\\
\bottomrule
\end{tabular}}
\vspace{-0.6em}
\caption{
Quantitative comparison on the \textit{FullBodyManipulation}~\cite{li2023object} with CHOIS~\cite{li2024controllable}, HOIFHLI~\cite{wu2025human}, and OMOMO~\cite{li2023object} variants (Lin-OMOMO and Pred-OMOMO) across four categories of evaluation metrics. 
Arrows indicate direction: ($\uparrow$) means higher is better, ($\downarrow$) means lower is better, and ($\rightarrow$) means closer to the real data value is better. 
The real-data $DIV$ reference is 9.02.
}
\label{tab:quantitative_FullBody}
\end{center}
\vspace{-2.4em}
\end{table*}
\begin{table}[t]
\begin{center}
\resizebox{\linewidth}{!}{%
\begin{tabular}{@{}lcccccccc@{}} 
\toprule
& \multicolumn{2}{c}{Condition Matching} 
& \multicolumn{4}{c}{Human Motion Quality} 
& \multicolumn{2}{c}{Interaction Quality} \\ 
\cmidrule(lr){2-3} \cmidrule(lr){4-7} \cmidrule(lr){8-9}
Methods & $T_s$$\downarrow$ & $T_e$$\downarrow$ & $H_{\text{feet}}$$\downarrow$ & $FS$$\downarrow$ & $R_{prec}$$\uparrow$ & $FID$$\downarrow$ & $C_{\%}$ & $P_{\text{hand}}$$\downarrow$ \\ 
\midrule
Lin-OMOMO~\cite{li2023object} & - & - & 7.92 & 0.48 & 0.57 & 8.85 & 0.26 & 0.20  \\
Pred-OMOMO~\cite{li2023object} & 4.72 & 10.92 & 6.61 & 0.45 & 0.59 & 5.16 & 0.40 & 0.17  \\
CHOIS~\cite{li2024controllable} & 5.75 & 10.28 & 4.20 & 0.42 & 0.61 & 2.04 & 0.46 & 0.18 \\
\midrule
\textbf{Ours (DecHOI)} & \textbf{4.27} & \textbf{8.43} & \textbf{4.06} & \textbf{0.41} & \textbf{0.69} & \textbf{1.01} & 0.48 & \textbf{0.15} \\
\bottomrule
\end{tabular}}
\vspace{-0.5em}
\caption{Quantitative results on the \textit{3D-FUTURE}~\cite{fu20213d}. DecHOI achieves better trajectory accuracy, motion stability, and contact realism than CHOIS~\cite{li2024controllable} and OMOMO~\cite{li2023object} baselines.
}
\label{tab:quantitative_3D_Future}
\end{center}
\vspace{-2.5em}
\end{table}

\subsection{Training and Inference}
\label{sec:Training}
\myparagraph{Objectives for Training.}
During optimization, the trajectory generator (TG) and the action generator (AG) are trained with distinct objectives. TG focuses exclusively on planning trajectories. Given the input $X$, it produces $\hat{\mathcal{T}}_0=\{\hat{\mathcal{T}}_o,\hat{\mathcal{T}}_h\}$ and is trained to reconstruct the clean trajectory representation $\mathcal{T}_0 = \{\mathcal{T}_o, \mathcal{T}_h\}$ with an $L_1 $ objective. The loss is defined as:
\vspace{-0.2em}
\begin{equation}
\label{eq:tg_loss}
\mathcal{L}_{\text{TG}} = \mathbb{E}_{\mathcal{T}_0, n \sim[1, N]} \| \hat{\mathcal{T}}_0 - \mathcal{T}_0 \|_1.
\end{equation}

AG receives the same input $X$ as TG, while the trajectories are kept clean. AG is trained to reconstruct the entire motion $P_0=\{P_o,P_h\}$ from which the generator predicts $\hat{P}_0=\{\hat{P}_o,\hat{P}_h\}$. The reconstruction objective is:
\vspace{-0.2em}
\begin{equation}
\label{eq:ag_loss}
\mathcal{L}_{\text{AG}} = \mathbb{E}_{P_0, n \sim[1, N]} \| \hat{P}_0 - P_0 \|_1.
\end{equation}

To further stabilize the reconstruction of the distal joints, we add a forward kinematic loss. Using the predicted relative joint rotations, we compute global hand positions $\hat{H}_0$ and global foot positions $\hat{F}_0$ from the pelvis root. These are supervised by clean distal joint positions $H_0$ and $F_0$ with:
\vspace{-0.2em}
\begin{equation}
\label{eq:fk_loss}
\mathcal{L}_{\text{FK}} = \|\hat{H}_0 - H_0\|_1 + \|\hat{F}_0 - F_0\|_1.
\end{equation}

The total objective for training the generators combines the reconstruction and adversarial terms:
\vspace{-0.2em}
\begin{equation}
\label{eq:total_loss}
\mathcal{L} = \lambda_{\text{TG}}\mathcal{L}_{\text{TG}} + \lambda_{\text{AG}}\mathcal{L}_{\text{AG}} + \lambda_{\text{FK}}\mathcal{L}_{\text{FK}} + \lambda_{\text{G}}\mathcal{L}_{\text{G}},
\end{equation}
where $\lambda_{\mathrm{TG}}$, $\lambda_{\mathrm{AG}}$, $\lambda_{\mathrm{FK}}$, and $\lambda_{\mathrm{G}}$ are scalar weights that balance the contributions of the terms.

\myparagraph{Inference-time Guidance.}
Inspired by previous work~\cite{li2024controllable}, our model applies reconstruction guidance to regularize the generation process. This design injects constraints at each denoising step without retraining the network for a specific purpose, which permits flexible control over the outputs. The process is formally represented as:
\vspace{-0.2em}
\begin{equation}
\label{eq:guidance}
\tilde{P}_0 = \hat{P}_0 - \alpha \Sigma_n \nabla_{P_n}\mathcal{F}(\hat{P}_0),
\end{equation}
where $\mathcal{F}$ is a regularization objective and $\alpha $ controls the perturbation strength. In our implementation, $\mathcal{F}$ encourages precise contact by penalizing distances between distal joints and the object surface, and stabilizes stance by minimizing deviations of foot joints from the ground plane to prevent hovering and penetration. Additional details are provided in the supplementary material.
\section{Experiments}
\label{sec:experiments}
\vspace{-.4em}
\subsection{Datasets}
Our experiments utilize two datasets designed to capture realistic human-object interactions across diverse indoor environments, following the evaluation setup in prior work~\cite{li2024controllable}.  
i) \textit{FullBodyManipulation}~\cite{li2023object}: Comprising about 10 hours of synchronized human-object motion sequences, this dataset covers 15 distinct rigid objects. 
We consider only rigid body interactions and therefore exclude sequences with articulated objects.
Motion data from 15 subjects are used for training, with data from two additional subjects reserved for testing, ensuring a consistent evaluation protocol.  
ii) \textit{3D-FUTURE}~\cite{fu20213d}: Containing a broad range of 3D furniture models. This dataset offers substantial geometric diversity. To evaluate generalization to unseen appearances and shapes, we substitute the 17 test objects in \textit{FullBodyManipulation} with unseen \textit{3D-FUTURE} models of the same categories, enabling systematic assessment on previously unobserved instances.


\subsection{Evaluation Metrics}

\paragraph{Condition Matching.}
This metric evaluates how accurately the generated motion aligns with the specified object conditions. We measure the Euclidean distance between the predicted and target object start and end positions $T_s$ and $T_e$ in the scene, reported in centimeters (cm).
\vspace{-1.3em}
\paragraph{Human Motion Quality.}
We evaluate realism and plausibility with five metrics: foot height ($H_{\text{feet}}$), foot sliding ($FS$), R-precision ($R_{\text{prec}}$), Fréchet Inception Distance ($FID$)~\cite{heusel2017gans}, and Diversity ($DIV$). $H_{\text{feet}}$ is the mean foot-to-ground height. $FS$ measures horizontal foot displacement during stance, following~\cite{li2024controllable}. Both are reported in centimeters. $R_{\text{prec}}$ (top-3) scores text-motion alignment. $FID$ compares generated and ground-truth motion distributions in a learned feature space. $DIV$ quantifies variation across samples under the same condition.
\vspace{-1.3em}
\paragraph{Interaction Quality.}
\begin{figure*}[t]
    \centering
    \includegraphics[width=1.0\textwidth]{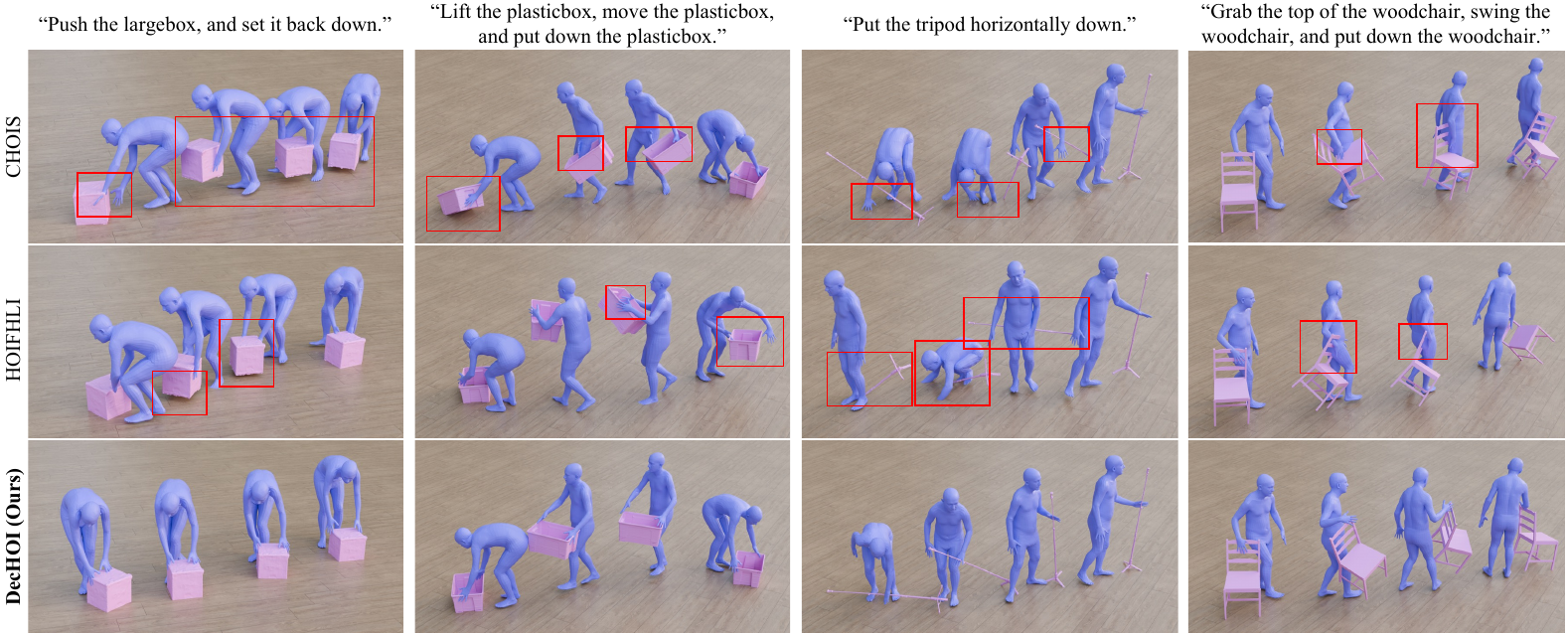}
    \vspace{-1.9em}
    \caption{
    Qualitative comparison of DecHOI with CHOIS~\cite{li2024controllable} and HOIFHLI~\cite{wu2025human} on the \textit{FullBodyManipulation}~\cite{li2023object}. DecHOI produces stable contacts, smooth motion, and accurate object trajectories, while prior methods show drift, penetration, or inconsistent coordination between human and object motions. 
    }
    \label{fig:qualitative}
    \vspace{-1.7em}
\end{figure*}
This category evaluates interaction accuracy and naturalness. We report hand-object contact precision ($C_{\text{prec}}$), recall ($C_{\text{rec}}$), F1 ($C_{F1}$), and percent ($C_{\%}$) as an overall contact reliability measure. We also report hand-object ($P_{\text{hand}}$) and body-object ($P_{\text{body}}$) penetration to quantify interpenetration depth ($cm$). 
\vspace{-1em}
\paragraph{Ground Truth Difference.}
This metric quantifies the deviation of generated motions from ground truth motions in both spatial and rotational domains. We report mean per-joint position error (MPJPE), root translation error ($T_{\text{root}}$), and object translation error ($T_{\text{obj}}$) as Euclidean distances in centimeters. Orientation consistency is evaluated with object orientation errors ($O_{\text{obj}}$), computed as the Frobenius norm of rotation differences.

\vspace{-0.3em}
\subsection{Quantitative Analysis}
\vspace{-0.3em}
We quantitatively evaluate our approach against CHOIS~\cite{li2024controllable}, HOIFHLI~\cite{wu2025human}, and two OMOMO~\cite{li2023object} variants (Lin- and Pred-OMOMO) implemented following CHOIS on two datasets: \textit{FullBodyManipulation}~\cite{li2023object} and \textit{3D-FUTURE}~\cite{fu20213d}. Because our method does not require specified intermediate waypoints, we evaluate all baselines without intermediate waypoint inputs to ensure a fair comparison. Note that Lin-OMOMO uses ground truth for all object-related signals, so metrics that are not applicable under this setting are reported as (–).
\vspace{-1.2em}
\paragraph{FullBodyManipulation.}
As shown in Tab.~\ref{tab:quantitative_FullBody}, our model surpasses all baselines on most metrics. OMOMO~\cite{li2023object} variants condition on object pose at every frame, yet under our extremely limited inputs they underperform in overall human motion and interaction quality. 
Methods relying on sparse object conditioning~\cite{li2024controllable, wu2025human} also degrade on contact-related measures once waypoints are removed, as the reduced conditioning increases task complexity and leads to unsynchronized interactions.
In contrast, DecHOI simplifies the problem through decoupled modeling, allowing the trajectory generator to produce accurate paths with low condition matching error and to attain competitive \textit{DIV} with reduced reliance on waypoints. The resulting interactions exhibit higher realism and stronger alignment with instructions, as reflected by improved \textit{FID} and $R_{prec}$. Although contact accuracy and penetration scores often trade off against each other, our adversarial training achieves balanced performance across both and produces realistic interactions. Further analyses and ablations are provided in the supplementary material.
\vspace{-1em}
\paragraph{3D-FUTURE.}
To further evaluate the generalization capability of our approach, we test it on unseen \textit{3D-FUTURE} samples that share the same object categories as those in the \textit{FullBodyManipulation} test set. As shown in Tab.~\ref{tab:quantitative_3D_Future}, our method maintains consistent trends across metrics, achieving strong trajectory accuracy and high contact reliability. Moreover, the improved $FID$ indicates better alignment between generated and GT motion distributions. These results demonstrate that DecHOI effectively synthesizes realistic and coherent human-object interactions even for previously unseen objects, highlighting its robust generalization and broad applicability across diverse objects.

\vspace{-.4em}
\subsection{Qualitative Analysis} 
\vspace{-0.3em}
In Fig.~\ref{fig:qualitative}, we present qualitative comparisons among CHOIS~\cite{li2024controllable}, HOIFHLI~\cite{wu2025human}, and our DecHOI on the \textit{FullBodyManipulation}~\cite{li2023object}. As illustrated in the four scenes, DecHOI consistently produces more stable and realistic interactions. 
For the \textit{largebox} scene, CHOIS and HOIFHLI yield unstable object motion, leading to box hovering and hand-object penetration. In contrast, DecHOI maintains a firmly grounded trajectory and cleaner contacts through stable global path modeling. 
In the \textit{plasticbox} scene, baselines exhibit severe penetration between the human body and the object, along with misaligned hand contacts. In contrast, our adversarial training on distal joints mitigates these failures and enhances contact realism. For the \textit{tripod} scene, both struggle with fine manipulation as the object drifts and rotates inconsistently, whereas our decoupled framework suppresses such unsynchronized artifacts and achieves more coherent motion. Lastly, in the \textit{woodchair} scene, prior works show noticeable mesh intersection and incomplete contact, while DecHOI produces stable contacts and precise spatial alignment between the hand and object.
\begin{figure}[t]
    \centering
    \includegraphics[width=\linewidth]{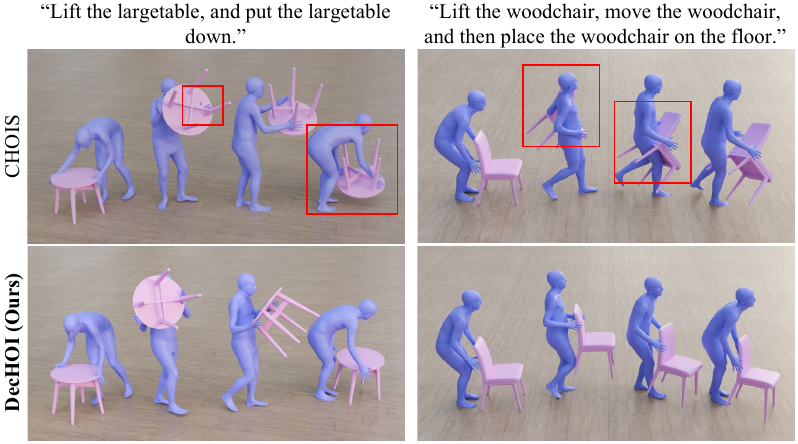}
    \vspace{-1.5em}
    \caption{
    Qualitative comparison of DecHOI and CHOIS~\cite{li2024controllable} on the \textit{3D-FUTURE}~\cite{fu20213d}, showing generalization to unseen objects.
    }
    \label{fig:3d_future}
    \vspace{-1.5em}
\end{figure}
As shown in Fig.~\ref{fig:3d_future}, we further compare CHOIS and DecHOI on two unseen objects from the \textit{3D-FUTURE} ~\cite{fu20213d}: \textit{largetable} and \textit{woodchair}. CHOIS often fails to adapt to these novel geometries, leading to human-object intersection and motion instability, whereas our method preserves accurate contacts and smooth motion, demonstrating strong generalization to unseen shapes. 
More qualitative results and analysis are provided in the supplementary material.

Additionally, Fig.~\ref{fig:loss_surface} visualizes the loss landscape to compare the training objective complexity of CHOIS, which uses a single network, and DecHOI. The surface of CHOIS appears noisy and highly rugged with many local minima, indicating sensitivity to initialization and unstable training. In contrast, DecHOI exhibits a smoother landscape with fewer local minima and more stable convergence. These observations suggest that our decoupled modeling effectively reduces overall optimization complexity.
\vspace{-0.5em}
\subsection{Long-term Dynamic Planning}
\vspace{-.2em}
To evaluate long-term and dynamic human-object interactions, we introduce \textit{DynaPlan}, which supports responsive planning and scene-aware interaction generation under multi-agent conditions. We equip both the agent and the moving counterpart with circular influence radii and detect potential collisions when their regions intersect. When a collision is detected, we re-plan with A*, adaptively choosing either a short detour or waiting to maintain goal-directed and collision-free motion while accounting for the future path of the counterpart. To handle uncertainty in that future path, we predict it with a pre-trained trajectory prediction network~\cite{mohamed2020social}. We evaluate performance on about 190 long-sequence indoor scenarios. Further details are provided in the supplementary material.
\vspace{-1em}
\paragraph{Quantitative Results.}
\begin{figure}[t]
    \centering
    \includegraphics[width=\linewidth]{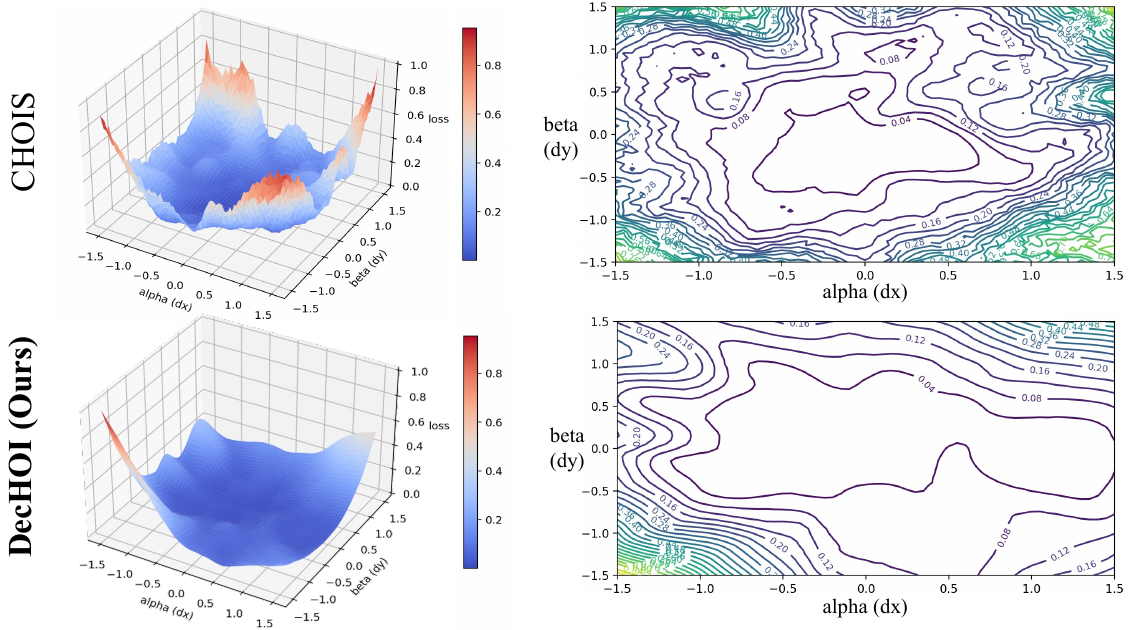}
    \vspace{-1.7em}
    \caption{
    Visualization of training loss landscapes for DecHOI and CHOIS~\cite{li2024controllable}, demonstrating reduced optimization complexity.
    }
    \label{fig:loss_surface}
    \vspace{-0.7em}
\end{figure}
\begin{table}[t]
\begin{center}
\resizebox{\linewidth}{!}{%
\begin{tabular}{@{}lcccccccccc@{}} 
\toprule
& \multicolumn{2}{c}{Condition Matching} 
& \multicolumn{2}{c}{Human Motion Quality} 
& \multicolumn{5}{c}{Interaction Quality} \\ 
\cmidrule(lr){2-3} \cmidrule(lr){4-5} \cmidrule(lr){6-10}
Methods & $T_s$$\downarrow$ & $T_e$$\downarrow$ & $H_{\text{feet}}$$\downarrow$ & $FS$$\downarrow$  & $C_{\%}$ & $P_{\text{hand}}$$\downarrow$ & $P_{\text{body}}$$\downarrow$ & $P_{\text{o$\rightarrow$}s}$$\downarrow$ & $P_{\text{h$\rightarrow$}s}$$\downarrow$ \\ 
\midrule
CHOIS~\cite{li2024controllable} & 2.19 & 8.05 & 3.14 & \textbf{0.43} & 0.78 & 0.73 & 0.73 & 1.00 & 0.83 \\
\midrule
\textbf{Ours (DecHOI)} & \textbf{1.90} & \textbf{7.98} & \textbf{2.85} & 0.45 & 0.76 & \textbf{0.65} & \textbf{0.63}  & \textbf{0.72} & \textbf{0.65} \\
\bottomrule
\end{tabular}}
\vspace{-0.5em}
\caption{
Quantitative results between CHOIS~\cite{li2024controllable} and DecHOI for responsive long-term interaction synthesis on the \textit{DynaPlan}.
}
\label{tab:quantitative_long_term}
\end{center}
\vspace{-2.5em}
\end{table}

Tab.~\ref{tab:quantitative_long_term} summarizes the quantitative results for long-sequence human-object interaction synthesis on the \textit{DynaPlan}. 
Compared to CHOIS~\cite{li2024controllable}, DecHOI achieves lower trajectory errors $T_s$ and $T_e$, along with reduced instability, resulting in smoother and more stable motion over extended sequences. 
For interaction quality, DecHOI shows clear improvements across all penetration metrics, including $P_{\text{hand}}$, $P_{\text{body}}$, $P_{\text{o}\rightarrow \text{s}}$, and $P_{\text{h}\rightarrow \text{s}}$. 
Here, $P_{\text{hand}}$ and $P_{\text{body}}$ measure the penetration depth between the human mesh and manipulated objects, while $P_{\text{o}\rightarrow \text{s}}$ and $P_{\text{h}\rightarrow \text{s}}$ quantify collisions between the object or human and the surrounding static scene. 
Lower values on most metrics indicate that DecHOI effectively minimizes interpenetration and maintains collision-free trajectories even in cluttered indoor environments. 

\begin{figure*}[t]
    \centering
    \includegraphics[width=1.0\textwidth]{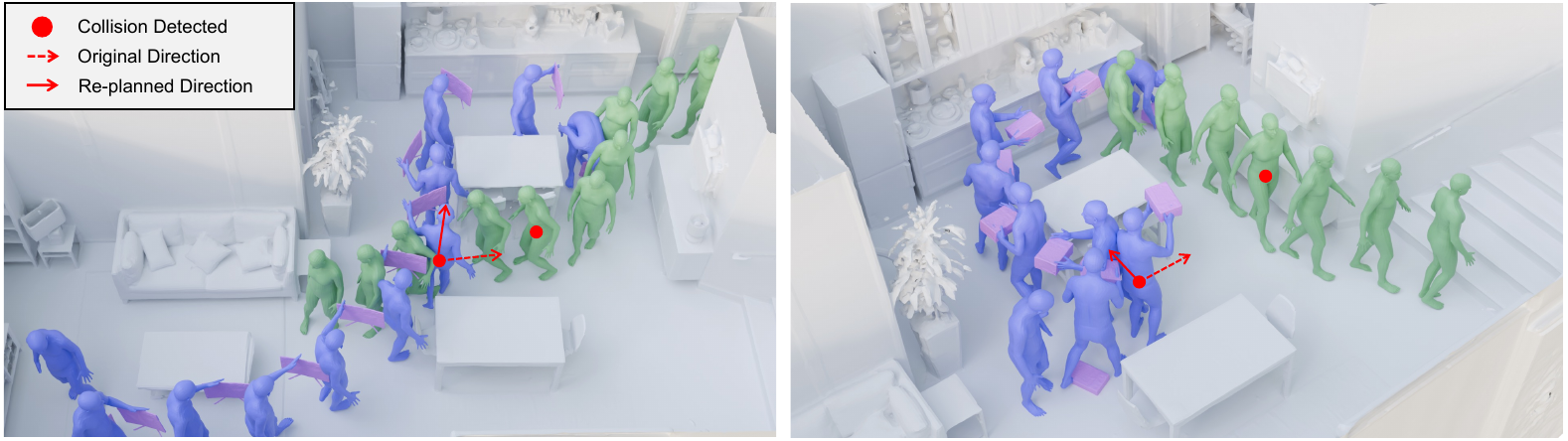}
    \vspace{-1.8em}
    \caption{
    Visualization of DecHOI in long-sequence dynamic environments. The human agent (blue) adaptively re-plans its path when encountering a moving obstacle (green), choosing between detour and waiting behaviors to maintain goal-directed, collision-free motion.
    }
    \label{fig:long}
    \vspace{-1.em}
\end{figure*}
These performance gains stem from the decoupled structure of the trajectory and action generators, which separates global path planning from detailed motion synthesis, combined with contact-aware adversarial learning that enforces spatial consistency throughout long-horizon interactions. 
Overall, DecHOI demonstrates superior stability, plausibility, and contact accuracy in dynamic multi-agent settings, outperforming CHOIS quantitatively.

\vspace{-1em}
\paragraph{Qualitative Results.}
Fig.~\ref{fig:long} visualizes DecHOI operating in two long-term dynamic indoor scenes from \textit{DynaPlan}. In both scenes, the agent encounters a moving counterpart along its original path and initiates reactive re-planning to maintain safe, goal-directed motion. Throughout these sequences, the agent maintains a continuous and valid trajectory with the manipulated object, and the action generator smoothly adapts joint motions to the updated trajectories. These results highlight DecHOI’s ability to sustain collision-free and consistent coordination between human and object across extended dynamic environments.

\subsection{Ablation Study}
\vspace{-.4em}
\begin{table}[t]
\begin{center}
\resizebox{\linewidth}{!}{%
\begin{tabular}{@{}lccccccc@{}} 
\toprule
& \multicolumn{2}{c}{Condition Matching} 
& \multicolumn{2}{c}{Human Motion Quality} 
& \multicolumn{3}{c}{Interaction Quality} \\ 
\cmidrule(lr){2-3} \cmidrule(lr){4-5} \cmidrule(lr){6-8}
Methods & $T_s$$\downarrow$ & $T_e$$\downarrow$ & $FS$$\downarrow$ & $R_{prec}$$\uparrow$ & $C_{F1}$$\uparrow$ & $P_{\text{hand}}$$\downarrow$ & $P_{\text{body}}$$\downarrow$ \\ 
\midrule
Baseline & 1.72 & 7.92 & \textbf{0.35} & 0.67 & 0.65 & 0.58 & 0.60 \\
w/ adversarial & 1.68 & 7.78 & 0.39 & 0.64 & 0.66 & \textbf{0.53} & 0.57 \\
w/ cross-attention  & 1.75 & 7.82 & 0.38 & 0.70 & 0.54 & 0.56 & 0.56 \\
\textbf{Ours (DecHOI)} & \textbf{1.59} & \textbf{6.91} & 0.38 & \textbf{0.72} & \textbf{0.67} & \textbf{0.53} & \textbf{0.54}\\
\bottomrule
\end{tabular}}
\vspace{-0.5em}
\caption{
Ablation results for DecHOI on the FullBodyManipulation~\cite{li2023object}, evaluating the contribution of each component.
}
\label{tab:ablation_1}
\end{center}
\vspace{-2.5em}
\end{table}

As shown in Tab.~\ref{tab:ablation_1}, we perform an ablation study on the \textit{FullBodyManipulation}~\cite{li2023object} to analyze the effects of adversarial training and text conditioning based on cross-attention. 
The baseline, which uses only the decoupled generative modeling and concatenates text embeddings along the sequence axis following prior work~\cite{li2024controllable, wu2025human}, achieves moderate performance but suffers from high penetration errors and limited contact consistency. 
Introducing adversarial training substantially reduces $P_{\text{hand}}$ and $P_{\text{body}}$, indicating improved realism and more stable interactions, although the linguistic alignment measured by $R_{\text{prec}}$ slightly decreases. 
Conversely, adding cross-attention layers for text conditioning enhances $R_{\text{prec}}$ by effectively capturing semantic intent, but slightly degrades interaction quality due to weaker regularization. 
When both modules are combined in DecHOI, the model achieves the best overall performance, simultaneously improving $R_{\text{prec}}$, $C_{F1}$, and penetration metrics. 
These results indicate that adversarial training enhances the robustness of distal joint interactions during object manipulation, and that text conditioning based on cross-attention enables more explicit transfer of semantic information across modalities, leading to stronger alignment.
\subsection{User Study}
\begin{figure}[t]
    \centering
    \vspace{-0.1em}
    \includegraphics[width=\linewidth]{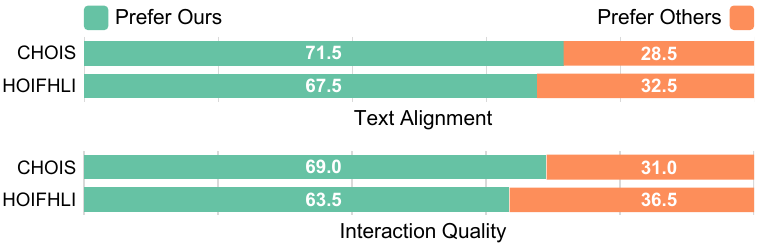}
    \vspace{-2em}
    \caption{
    Stacked horizontal bars showing user preference distributions for DecHOI compared with CHOIS~\cite{li2024controllable} and HOIFHLI~\cite{wu2025human} across two evaluation criteria: Text Alignment and Interaction Quality.
    }
    \label{fig:user}
    \vspace{-1.5em}
\end{figure}
\vspace{-0.5em}
We conducted a user study to evaluate the perceptual realism and text alignment of DecHOI compared with CHOIS~\cite{li2024controllable} and HOIFHLI~\cite{wu2025human}. 
For each comparison, a total of 60 text-scene pairs, 30 per method, were randomly sampled from our full test set, and 200 participants were recruited via Amazon Mechanical Turk (AMT).
Each participant viewed two anonymized clips, each lasting about four seconds, in random order and answered two questions: i)~which video better matches the text instruction (Text Alignment), and ii)~which video shows better human-object interaction quality (Interaction Quality). Responses were collected using a two-choice scale: \textit{Prefer Ours} or \textit{Prefer Others}. As summarized in Fig.~\ref{fig:user}, DecHOI was consistently preferred on both criteria across both baselines, confirming that our decoupled framework yields motions that are semantically faithful to textual instructions and more realistic, stable, and natural than prior methods.

\section{Conclusion}
\vspace{-.2em}
\label{sec:conclusion}

In this work, we propose a decoupled generative modeling framework for human-object interaction synthesis that separates trajectory generation from fine-grained action generation. This formulation lowers optimization complexity and enables the model to learn precise and stable motion, which reduces temporal asynchrony and penetration. 
It also removes the need for manually specified intermediate waypoints, improving practical applicability.
Additionally, a compact adversarial discriminator focused on distal joint cues further improves interaction fidelity and strengthens contact realism. We also present a dynamic planner that delivers robust and adaptive path planning in long-horizon, multi-agent settings, as demonstrated on the \textit{DynaPlan}. Extensive quantitative and qualitative results show state-of-the-art performance across major metrics, producing realistic and semantically aligned human-object interaction.

\section*{Acknowledgement}
This work was supported by Culture, Sports and Tourism R\&D Program through the Korea Creative Content Agency grant funded by the Ministry of Culture, Sports and Tourism 
(International Collaborative Research and Global Talent Development for the Development of Copyright Management and Protection Technologies for Generative AI, RS-2024-00345025, 46\%),
the National Research Foundation of Korea(NRF) grant funded by the Korea government(MSIT)(RS-2025-00521602, 50\%),
Institute of Information \& communications Technology Planning \& Evaluation (IITP) \& ITRC(Information Technology Research Center) grant funded by the Korea government(MSIT) (No.RS-2019-II190079, Artificial Intelligence Graduate School Program(Korea University), 1\%;
IITP-2025-RS-2024-00436857, 1\%), 
the National Science Foundation, United States under Grant (NSF) Nos. IIS-2413161 and IIS-2504906,
the NVIDIA Academic Grant Program using NVIDIA RTX PRO 6000 Max-Q Workstation Edition.

{
    \small
    \bibliographystyle{ieeenat_fullname}
    \bibliography{main}

@String(TOG= {ACM Trans. Graph.})

@String(ICLR = {Int. Conf. Learn. Represent.})

@String(AAAI = {AAAI})

@String(TOG   = {ACM TOG})

@String(ICLR  = {ICLR})

@inproceedings{li2024controllable,
  title={Controllable human-object interaction synthesis},
  author={Li, Jiaman and Clegg, Alexander and Mottaghi, Roozbeh and Wu, Jiajun and Puig, Xavier and Liu, C Karen},
  booktitle={European Conference on Computer Vision},
  pages={54--72},
  year={2024},
  organization={Springer}
}

@inproceedings{wu2025human,
  title={Human-object interaction from human-level instructions},
  author={Wu, Zhen and Li, Jiaman and Xu, Pei and Liu, C Karen},
  booktitle={Proceedings of the IEEE/CVF International Conference on Computer Vision},
  pages={11176--11186},
  year={2025}
}

@article{li2023object,
  title={Object motion guided human motion synthesis},
  author={Li, Jiaman and Wu, Jiajun and Liu, C Karen},
  journal={ACM Transactions on Graphics (TOG)},
  volume={42},
  number={6},
  pages={1--11},
  year={2023},
  publisher={ACM New York, NY, USA}
}

@article{fu20213d,
  title={3d-future: 3d furniture shape with texture},
  author={Fu, Huan and Jia, Rongfei and Gao, Lin and Gong, Mingming and Zhao, Binqiang and Maybank, Steve and Tao, Dacheng},
  journal={International Journal of Computer Vision},
  volume={129},
  number={12},
  pages={3313--3337},
  year={2021},
  publisher={Springer}
}

@inproceedings{prokudin2019efficient,
  title={Efficient learning on point clouds with basis point sets},
  author={Prokudin, Sergey and Lassner, Christoph and Romero, Javier},
  booktitle={Proceedings of the IEEE/CVF international conference on computer vision},
  pages={4332--4341},
  year={2019}
}

@article{ho2020denoising,
  title={Denoising diffusion probabilistic models},
  author={Ho, Jonathan and Jain, Ajay and Abbeel, Pieter},
  journal={Advances in neural information processing systems},
  volume={33},
  pages={6840--6851},
  year={2020}
}

@inproceedings{radford2021learning,
  title={Learning transferable visual models from natural language supervision},
  author={Radford, Alec and Kim, Jong Wook and Hallacy, Chris and Ramesh, Aditya and Goh, Gabriel and Agarwal, Sandhini and Sastry, Girish and Askell, Amanda and Mishkin, Pamela and Clark, Jack and others},
  booktitle={International conference on machine learning},
  pages={8748--8763},
  year={2021},
  organization={PmLR}
}

@inproceedings{guo2022generating,
  title={Generating diverse and natural 3d human motions from text},
  author={Guo, Chuan and Zou, Shihao and Zuo, Xinxin and Wang, Sen and Ji, Wei and Li, Xingyu and Cheng, Li},
  booktitle={Proceedings of the IEEE/CVF conference on computer vision and pattern recognition},
  pages={5152--5161},
  year={2022}
}

@inproceedings{
tevet2023human,
title={Human Motion Diffusion Model},
author={Guy Tevet and Sigal Raab and Brian Gordon and Yoni Shafir and Daniel Cohen-or and Amit Haim Bermano},
booktitle={The Eleventh International Conference on Learning Representations },
year={2023},
url={https://openreview.net/forum?id=SJ1kSyO2jwu}
}

@inproceedings{shafir2024human,
  title={Human Motion Diffusion as a Generative Prior},
  author={Shafir, Yoni and Tevet, Guy and Kapon, Roy and Bermano, Amit Haim},
  year={2024},
  booktitle={The Twelfth International Conference on Learning Representations}
}

@article{meng2025absolute,
  title={Absolute Coordinates Make Motion Generation Easy},
  author={Meng, Zichong and Han, Zeyu and Peng, Xiaogang and Xie, Yiming and Jiang, Huaizu},
  journal={arXiv preprint arXiv:2505.19377},
  year={2025}
}

@inproceedings{
xie2024omnicontrol,
title={OmniControl: Control Any Joint at Any Time for Human Motion Generation},
author={Yiming Xie and Varun Jampani and Lei Zhong and Deqing Sun and Huaizu Jiang},
booktitle={The Twelfth International Conference on Learning Representations},
year={2024},
url={https://openreview.net/forum?id=gd0lAEtWso}
}

@inproceedings{guo2024momask,
  title={Momask: Generative masked modeling of 3d human motions},
  author={Guo, Chuan and Mu, Yuxuan and Javed, Muhammad Gohar and Wang, Sen and Cheng, Li},
  booktitle={Proceedings of the IEEE/CVF Conference on Computer Vision and Pattern Recognition},
  pages={1900--1910},
  year={2024}
}

@inproceedings{diller2024cg,
  title={Cg-hoi: Contact-guided 3d human-object interaction generation},
  author={Diller, Christian and Dai, Angela},
  booktitle={Proceedings of the IEEE/CVF Conference on Computer Vision and Pattern Recognition},
  pages={19888--19901},
  year={2024}
}

@inproceedings{kulkarni2024nifty,
  title={Nifty: Neural object interaction fields for guided human motion synthesis},
  author={Kulkarni, Nilesh and Rempe, Davis and Genova, Kyle and Kundu, Abhijit and Johnson, Justin and Fouhey, David and Guibas, Leonidas},
  booktitle={Proceedings of the IEEE/CVF Conference on Computer Vision and Pattern Recognition},
  pages={947--957},
  year={2024}
}

@article{ho2022classifier,
  title={Classifier-free diffusion guidance},
  author={Ho, Jonathan and Salimans, Tim},
  journal={arXiv preprint arXiv:2207.12598},
  year={2022}
}

@inproceedings{pavlakos2019expressive,
  title={Expressive body capture: 3d hands, face, and body from a single image},
  author={Pavlakos, Georgios and Choutas, Vasileios and Ghorbani, Nima and Bolkart, Timo and Osman, Ahmed AA and Tzionas, Dimitrios and Black, Michael J},
  booktitle={Proceedings of the IEEE/CVF conference on computer vision and pattern recognition},
  pages={10975--10985},
  year={2019}
}

@inproceedings{mohamed2020social,
  title={Social-stgcnn: A social spatio-temporal graph convolutional neural network for human trajectory prediction},
  author={Mohamed, Abduallah and Qian, Kun and Elhoseiny, Mohamed and Claudel, Christian},
  booktitle={Proceedings of the IEEE/CVF conference on computer vision and pattern recognition},
  pages={14424--14432},
  year={2020}
}

@article{straub2019replica,
  title={The replica dataset: A digital replica of indoor spaces},
  author={Straub, Julian and Whelan, Thomas and Ma, Lingni and Chen, Yufan and Wijmans, Erik and Green, Simon and Engel, Jakob J and Mur-Artal, Raul and Ren, Carl and Verma, Shobhit and others},
  journal={arXiv preprint arXiv:1906.05797},
  year={2019}
}

@article{heusel2017gans,
  title={Gans trained by a two time-scale update rule converge to a local nash equilibrium},
  author={Heusel, Martin and Ramsauer, Hubert and Unterthiner, Thomas and Nessler, Bernhard and Hochreiter, Sepp},
  journal={Advances in neural information processing systems},
  volume={30},
  year={2017}
}

@inproceedings{bhatnagar2022behave,
  title={Behave: Dataset and method for tracking human object interactions},
  author={Bhatnagar, Bharat Lal and Xie, Xianghui and Petrov, Ilya A and Sminchisescu, Cristian and Theobalt, Christian and Pons-Moll, Gerard},
  booktitle={Proceedings of the IEEE/CVF Conference on Computer Vision and Pattern Recognition},
  pages={15935--15946},
  year={2022}
}

@inproceedings{xu2025interact,
  title={Interact: Advancing large-scale versatile 3d human-object interaction generation},
  author={Xu, Sirui and Li, Dongting and Zhang, Yucheng and Xu, Xiyan and Long, Qi and Wang, Ziyin and Lu, Yunzhi and Dong, Shuchang and Jiang, Hezi and Gupta, Akshat and others},
  booktitle={Proceedings of the Computer Vision and Pattern Recognition Conference},
  pages={7048--7060},
  year={2025}
}

@inproceedings{liu2022hoi4d,
  title={Hoi4d: A 4d egocentric dataset for category-level human-object interaction},
  author={Liu, Yunze and Liu, Yun and Jiang, Che and Lyu, Kangbo and Wan, Weikang and Shen, Hao and Liang, Boqiang and Fu, Zhoujie and Wang, He and Yi, Li},
  booktitle={Proceedings of the IEEE/CVF Conference on Computer Vision and Pattern Recognition},
  pages={21013--21022},
  year={2022}
}

@article{zhu2023human,
  title={Human motion generation: A survey},
  author={Zhu, Wentao and Ma, Xiaoxuan and Ro, Dongwoo and Ci, Hai and Zhang, Jinlu and Shi, Jiaxin and Gao, Feng and Tian, Qi and Wang, Yizhou},
  journal={IEEE Transactions on Pattern Analysis and Machine Intelligence},
  volume={46},
  number={4},
  pages={2430--2449},
  year={2023},
  publisher={IEEE}
}

@article{xue2025human,
  title={Human motion video generation: A survey},
  author={Xue, Haiwei and Luo, Xiangyang and Hu, Zhanghao and Zhang, Xin and Xiang, Xunzhi and Dai, Yuqin and Liu, Jianzhuang and Zhang, Zhensong and Li, Minglei and Yang, Jian and others},
  journal={IEEE Transactions on Pattern Analysis and Machine Intelligence},
  year={2025},
  publisher={IEEE}
}

@article{antoun2023human,
  title={Human object interaction detection: Design and survey},
  author={Antoun, Maya and Asmar, Daniel},
  journal={Image and Vision Computing},
  volume={130},
  pages={104617},
  year={2023},
  publisher={Elsevier}
}

@inproceedings{hao2024hand,
  title={Hand-centric motion refinement for 3d hand-object interaction via hierarchical spatial-temporal modeling},
  author={Hao, Yuze and Zhang, Jianrong and Zhuo, Tao and Wen, Fuan and Fan, Hehe},
  booktitle={Proceedings of the AAAI Conference on Artificial Intelligence},
  volume={38},
  number={3},
  pages={2076--2084},
  year={2024}
}

@article{mao2022contact,
  title={Contact-aware human motion forecasting},
  author={Mao, Wei and Hartley, Richard I and Salzmann, Mathieu and others},
  journal={Advances in Neural Information Processing Systems},
  volume={35},
  pages={7356--7367},
  year={2022}
}

@inproceedings{cseke2025pico,
  title={PICO: Reconstructing 3D people in contact with objects},
  author={Cseke, Alp{\'a}r and Tripathi, Shashank and Dwivedi, Sai Kumar and Lakshmipathy, Arjun S and Chatterjee, Agniv and Black, Michael J and Tzionas, Dimitrios},
  booktitle={Proceedings of the Computer Vision and Pattern Recognition Conference},
  pages={1783--1794},
  year={2025}
}

@inproceedings{gu2024contactgen,
  title={Contactgen: Contact-guided interactive 3d human generation for partners},
  author={Gu, Dongjun and Shim, Jaehyeok and Jang, Jaehoon and Kang, Changwoo and Joo, Kyungdon},
  booktitle={Proceedings of the AAAI Conference on Artificial Intelligence},
  volume={38},
  number={3},
  pages={1923--1931},
  year={2024}
}

@inproceedings{yuan2023physdiff,
  title={Physdiff: Physics-guided human motion diffusion model},
  author={Yuan, Ye and Song, Jiaming and Iqbal, Umar and Vahdat, Arash and Kautz, Jan},
  booktitle={Proceedings of the IEEE/CVF international conference on computer vision},
  pages={16010--16021},
  year={2023}
}

@inproceedings{lee2023locomotion,
  title={Locomotion-action-manipulation: Synthesizing human-scene interactions in complex 3d environments},
  author={Lee, Jiye and Joo, Hanbyul},
  booktitle={Proceedings of the IEEE/CVF International Conference on Computer Vision},
  pages={9663--9674},
  year={2023}
}

@article{jiang2023motiongpt,
  title={Motiongpt: Human motion as a foreign language},
  author={Jiang, Biao and Chen, Xin and Liu, Wen and Yu, Jingyi and Yu, Gang and Chen, Tao},
  journal={Advances in Neural Information Processing Systems},
  volume={36},
  pages={20067--20079},
  year={2023}
}

@inproceedings{meng2025rethinking,
  title={Rethinking Diffusion for Text-Driven Human Motion Generation: Redundant Representations, Evaluation, and Masked Autoregression},
  author={Meng, Zichong and Xie, Yiming and Peng, Xiaogang and Han, Zeyu and Jiang, Huaizu},
  booktitle={Proceedings of the Computer Vision and Pattern Recognition Conference},
  pages={27859--27871},
  year={2025}
}

@inproceedings{fan2024freemotion,
  title={Freemotion: A unified framework for number-free text-to-motion synthesis},
  author={Fan, Ke and Tang, Junshu and Cao, Weijian and Yi, Ran and Li, Moran and Gong, Jingyu and Zhang, Jiangning and Wang, Yabiao and Wang, Chengjie and Ma, Lizhuang},
  booktitle={European Conference on Computer Vision},
  pages={93--109},
  year={2024},
  organization={Springer}
}

@inproceedings{zou2024parco,
  title={Parco: Part-coordinating text-to-motion synthesis},
  author={Zou, Qiran and Yuan, Shangyuan and Du, Shian and Wang, Yu and Liu, Chang and Xu, Yi and Chen, Jie and Ji, Xiangyang},
  booktitle={European Conference on Computer Vision},
  pages={126--143},
  year={2024},
  organization={Springer}
}

@inproceedings{xue2025guiding,
  title={Guiding Human-Object Interactions with Rich Geometry and Relations},
  author={Xue, Mengqing and Liu, Yifei and Guo, Ling and Huang, Shaoli and Ding, Changxing},
  booktitle={Proceedings of the Computer Vision and Pattern Recognition Conference},
  pages={22714--22723},
  year={2025}
}

@inproceedings{jia2025primhoi,
  title={PrimHOI: Compositional Human-Object Interaction via Reusable Primitives},
  author={Jia, Kai and Liu, Tengyu and Pei, Mingtao and Zhu, Yixin and Huang, Siyuan},
  booktitle={Proceedings of the IEEE/CVF International Conference on Computer Vision},
  pages={11491--11501},
  year={2025}
}

@inproceedings{zhang2025interactanything,
  title={InteractAnything: Zero-shot Human Object Interaction Synthesis via LLM Feedback and Object Affordance Parsing},
  author={Zhang, Jinlu and Chen, Yixin and Wang, Zan and Yang, Jie and Wang, Yizhou and Huang, Siyuan},
  booktitle={Proceedings of the Computer Vision and Pattern Recognition Conference},
  pages={7015--7025},
  year={2025}
}

@inproceedings{song2024hoianimator,
  title={Hoianimator: Generating text-prompt human-object animations using novel perceptive diffusion models},
  author={Song, Wenfeng and Zhang, Xinyu and Li, Shuai and Gao, Yang and Hao, Aimin and Hou, Xia and Chen, Chenglizhao and Li, Ning and Qin, Hong},
  booktitle={Proceedings of the IEEE/CVF Conference on Computer Vision and Pattern Recognition},
  pages={811--820},
  year={2024}
}

@inproceedings{cen2024generating,
  title={Generating human motion in 3d scenes from text descriptions},
  author={Cen, Zhi and Pi, Huaijin and Peng, Sida and Shen, Zehong and Yang, Minghui and Zhu, Shuai and Bao, Hujun and Zhou, Xiaowei},
  booktitle={Proceedings of the IEEE/CVF conference on computer vision and pattern recognition},
  pages={1855--1866},
  year={2024}
}

@inproceedings{zhang2025energymogen,
  title={Energymogen: Compositional human motion generation with energy-based diffusion model in latent space},
  author={Zhang, Jianrong and Fan, Hehe and Yang, Yi},
  booktitle={Proceedings of the Computer Vision and Pattern Recognition Conference},
  pages={17592--17602},
  year={2025}
}

@inproceedings{yi2024generating,
  title={Generating human interaction motions in scenes with text control},
  author={Yi, Hongwei and Thies, Justus and Black, Michael J and Peng, Xue Bin and Rempe, Davis},
  booktitle={European Conference on Computer Vision},
  pages={246--263},
  year={2024},
  organization={Springer}
}

@inproceedings{wang2025hsi,
  title={HSI-GPT: A General-Purpose Large Scene-Motion-Language Model for Human Scene Interaction},
  author={Wang, Yuan and Li, Yali and Li, Xiang and Wang, Shengjin},
  booktitle={Proceedings of the Computer Vision and Pattern Recognition Conference},
  pages={7147--7157},
  year={2025}
}

@inproceedings{wang2024move,
  title={Move as you say interact as you can: Language-guided human motion generation with scene affordance},
  author={Wang, Zan and Chen, Yixin and Jia, Baoxiong and Li, Puhao and Zhang, Jinlu and Zhang, Jingze and Liu, Tengyu and Zhu, Yixin and Liang, Wei and Huang, Siyuan},
  booktitle={Proceedings of the IEEE/CVF Conference on Computer Vision and Pattern Recognition},
  pages={433--444},
  year={2024}
}

@inproceedings{li2024genzi,
  title={Genzi: Zero-shot 3d human-scene interaction generation},
  author={Li, Lei and Dai, Angela},
  booktitle={Proceedings of the IEEE/CVF Conference on Computer Vision and Pattern Recognition},
  pages={20465--20474},
  year={2024}
}

@inproceedings{yang2024f,
  title={F-hoi: Toward fine-grained semantic-aligned 3d human-object interactions},
  author={Yang, Jie and Niu, Xuesong and Jiang, Nan and Zhang, Ruimao and Huang, Siyuan},
  booktitle={European Conference on Computer Vision},
  pages={91--110},
  year={2024},
  organization={Springer}
}

@article{cao2023detecting,
  title={Detecting any human-object interaction relationship: Universal hoi detector with spatial prompt learning on foundation models},
  author={Cao, Yichao and Tang, Qingfei and Su, Xiu and Chen, Song and You, Shan and Lu, Xiaobo and Xu, Chang},
  journal={Advances in Neural Information Processing Systems},
  volume={36},
  pages={739--751},
  year={2023}
}

@inproceedings{chen2023detecting,
  title={Detecting human-object contact in images},
  author={Chen, Yixin and Dwivedi, Sai Kumar and Black, Michael J and Tzionas, Dimitrios},
  booktitle={Proceedings of the IEEE/CVF Conference on Computer Vision and Pattern Recognition},
  pages={17100--17110},
  year={2023}
}

@inproceedings{liu2025core4d,
  title={Core4d: A 4d human-object-human interaction dataset for collaborative object rearrangement},
  author={Liu, Yun and Zhang, Chengwen and Xing, Ruofan and Tang, Bingda and Yang, Bowen and Yi, Li},
  booktitle={Proceedings of the Computer Vision and Pattern Recognition Conference},
  pages={1769--1782},
  year={2025}
}

@inproceedings{yang2024lemon,
  title={Lemon: Learning 3d human-object interaction relation from 2d images},
  author={Yang, Yuhang and Zhai, Wei and Luo, Hongchen and Cao, Yang and Zha, Zheng-Jun},
  booktitle={Proceedings of the IEEE/CVF conference on computer vision and pattern recognition},
  pages={16284--16295},
  year={2024}
}

@inproceedings{xu2025intermimic,
  title={Intermimic: Towards universal whole-body control for physics-based human-object interactions},
  author={Xu, Sirui and Ling, Hung Yu and Wang, Yu-Xiong and Gui, Liang-Yan},
  booktitle={Proceedings of the Computer Vision and Pattern Recognition Conference},
  pages={12266--12277},
  year={2025}
}

@inproceedings{jiang2024scaling,
  title={Scaling up dynamic human-scene interaction modeling},
  author={Jiang, Nan and Zhang, Zhiyuan and Li, Hongjie and Ma, Xiaoxuan and Wang, Zan and Chen, Yixin and Liu, Tengyu and Zhu, Yixin and Huang, Siyuan},
  booktitle={Proceedings of the IEEE/CVF Conference on Computer Vision and Pattern Recognition},
  pages={1737--1747},
  year={2024}
}

@inproceedings{huang2022capturing,
  title={Capturing and inferring dense full-body human-scene contact},
  author={Huang, Chun-Hao P and Yi, Hongwei and H{\"o}schle, Markus and Safroshkin, Matvey and Alexiadis, Tsvetelina and Polikovsky, Senya and Scharstein, Daniel and Black, Michael J},
  booktitle={Proceedings of the IEEE/CVF Conference on Computer Vision and Pattern Recognition},
  pages={13274--13285},
  year={2022}
}

@inproceedings{tripathi2023deco,
  title={Deco: Dense estimation of 3d human-scene contact in the wild},
  author={Tripathi, Shashank and Chatterjee, Agniv and Passy, Jean-Claude and Yi, Hongwei and Tzionas, Dimitrios and Black, Michael J},
  booktitle={Proceedings of the IEEE/CVF international conference on computer vision},
  pages={8001--8013},
  year={2023}
}

@inproceedings{xu2023interdiff,
  title={Interdiff: Generating 3d human-object interactions with physics-informed diffusion},
  author={Xu, Sirui and Li, Zhengyuan and Wang, Yu-Xiong and Gui, Liang-Yan},
  booktitle={Proceedings of the IEEE/CVF International Conference on Computer Vision},
  pages={14928--14940},
  year={2023}
}

@inproceedings{savva2019habitat,
  title={Habitat: A platform for embodied ai research},
  author={Savva, Manolis and Kadian, Abhishek and Maksymets, Oleksandr and Zhao, Yili and Wijmans, Erik and Jain, Bhavana and Straub, Julian and Liu, Jia and Koltun, Vladlen and Malik, Jitendra and others},
  booktitle={Proceedings of the IEEE/CVF international conference on computer vision},
  pages={9339--9347},
  year={2019}
}

@article{szot2021habitat,
  title={Habitat 2.0: Training home assistants to rearrange their habitat},
  author={Szot, Andrew and Clegg, Alexander and Undersander, Eric and Wijmans, Erik and Zhao, Yili and Turner, John and Maestre, Noah and Mukadam, Mustafa and Chaplot, Devendra Singh and Maksymets, Oleksandr and others},
  journal={Advances in neural information processing systems},
  volume={34},
  pages={251--266},
  year={2021}
}

@article{li2018visualizing,
  title={Visualizing the loss landscape of neural nets},
  author={Li, Hao and Xu, Zheng and Taylor, Gavin and Studer, Christoph and Goldstein, Tom},
  journal={Advances in neural information processing systems},
  volume={31},
  year={2018}
}

@inproceedings{kingma2015adam,
  author    = {Kingma, Diederik P. and Ba, Jimmy},
  title     = {Adam: A Method for Stochastic Optimization},
  booktitle = {International Conference on Learning Representations (ICLR)},
  year      = {2015},
  url       = {https://arxiv.org/abs/1412.6980}
}

@article{williams1989learning,
  title={A learning algorithm for continually running fully recurrent neural networks},
  author={Williams, Ronald J and Zipser, David},
  journal={Neural computation},
  volume={1},
  number={2},
  pages={270--280},
  year={1989},
  publisher={MIT Press One Rogers Street, Cambridge, MA 02142-1209, USA journals-info~…}
}

@inproceedings{martinez2017simple,
  title={A simple yet effective baseline for 3d human pose estimation},
  author={Martinez, Julieta and Hossain, Rayat and Romero, Javier and Little, James J},
  booktitle={Proceedings of the IEEE international conference on computer vision},
  pages={2640--2649},
  year={2017}
}
}


\end{document}



\renewcommand{\thesection}{\Alph{section}} 
\newcommand{\myparagraph}[1]{\vspace{4pt}\noindent{\bf #1}}
\newcommand{\secbest}[1]{#1}

\title{
Decoupled Generative Modeling for Human-Object Interaction Synthesis}

\thispagestyle{empty}
\maketitlesupplementary


\section*{Overview}
\noindent In this supplementary material, we provide additional details and analyses to complement the main paper.
Section~\ref{sec_a} summarizes the experimental setup, including key implementation details and training and inference configurations.
Section~\ref{sec_b} defines the evaluation protocols and metrics, while 
Section~\ref{sec_c} introduces our reconstruction guidance for improving hand-object contact and foot grounding.
Section~\ref{sec_d} describes DynaPlan for collision-aware dynamic planning, and
Section~\ref{sec_e} reports additional experiments, including an ablation on the generator loss weight, comparisons under privileged waypoint supervision, and an oracle trajectory study.
Sections~\ref{sec_f} to ~\ref{sec_j} detail the loss landscape visualization, the implementation of the OMOMO variants, the user study design, extended qualitative results, and limitations and future directions.
\section{Experimental Setup}
\label{sec_a}
\subsection{Implementation Details}
\label{supp_sec:impl_details}
Our models are implemented in the PyTorch deep learning framework. All experiments are conducted on a single NVIDIA RTX 3090 GPU with 24 GB of memory, and training requires approximately 25 GPU hours. We optimize the generators with the Adam optimizer~\cite{kingma2015adam} using a learning rate of $1 \times 10^{-4}$ and a batch size of 32, where each training sample is a sequence with $T = 120$ frames. During training, input sequences shorter than $T$ are zero-padded to match the length. The transformer-based denoising network consists of 4 attention heads and 4 layers. We train the model for 580k steps and report results using the checkpoint that achieves the best validation performance. The diffusion process uses 1,000 noising steps during training, and we employ the standard DDPM~\cite{ho2020denoising} sampling procedure at inference time.

For adversarial training of the discriminator, we use the same optimizer configuration as for the generators, but set the learning rate to $2 \times 10^{-5}$. The discriminator is updated once for every generator update. The loss balancing weights for the trajectory generator, action generator, and forward kinematics objectives are fixed to $\lambda_{\text{TG}} = 0.1$, $\lambda_{\text{AG}} = 1.0$, and $\lambda_{\text{FK}} = 1.0$, respectively. The adversarial loss weight $\lambda_{\text{G}}$ is adjusted within the range [$0.01, 0.05]$ during training.

In addition, to prevent error accumulation from the trajectory generator (TG) propagating into the action generator (AG), we provide clean trajectories for both the human and the object as part of the noisy input to AG during training. In other words, AG is conditioned on ground-truth trajectories while the remaining components are corrupted with noise, which is inspired by the teacher forcing strategy~\cite{williams1989learning} commonly used to mitigate error accumulation. 
Note that during training TG receives noisy inputs for all frames except the start and goal conditions. At inference time the trajectories predicted by TG are then used as inputs to AG.

\subsection{Inference Runtime and Resource Consumption}
\label{supp_sec:inference_time}
\begin{table}[h]
\begin{center}
\scalebox{0.75}{%
\begin{tabular}{@{}lcc@{}} \toprule
    Methods & Inference time (s) & Inference VRAM (GB) \\ \midrule
    CHOIS~\cite{li2024controllable} &  2.00 & 1.9  \\
    HOIFHLI~\cite{wu2025human} & 162.71 & 5.2 \\
    Ours (DecHOI) & 3.41 & 2.1 \\
\bottomrule
\end{tabular}}
\caption{
Inference runtime and GPU memory usage for CHOIS~\cite{li2024controllable}, HOIFHLI~\cite{wu2025human}, and our DecHOI.
}
\end{center}
\end{table}
In this section, we report the computational cost of DecHOI and the baselines used in our comparison. DecHOI employs two lightweight denoising networks as specialized experts for the decoupled generative modeling of trajectories and interactions. This design incurs a modest increase in inference time compared to CHOIS~\cite{li2024controllable}, while remaining significantly more efficient than HOIFHLI~\cite{wu2025human}, which relies on a generation pipeline with multi-stage for grasp generation and refinement and therefore has substantially higher inference time.

We further observe that DecHOI requires comparable or even lower VRAM usage than the prior models. Considering the qualitative and quantitative improvements, the additional inference cost of DecHOI is modest relative to the overall computational budget. Moreover, when taking into account the manual effort required to annotate intermediate waypoints for the baselines, DecHOI offers a more practical solution in realistic deployment scenarios.

\section{Evaluation Details}
\label{sec_b}
For an accurate and fair comparison, all trajectories are expressed in a common global world frame, and all protocols and thresholds used for metric computation follow the same settings as CHOIS~\cite{li2024controllable}.
\subsection{Condition Matching}
We measure $T_s$ and $T_e$ as the Euclidean distances between the predicted and ground-truth object centroids at the start frame and at the final target position, respectively.
Since our generator uses only the start and goal points, we evaluate condition matching solely with $T_s$ and $T_e$.

\subsection{Human Motion Quality}
\myparagraph{Foot height.}
$H_{\text{feet}}$ is the mean distance from the feet to the floor.
To obtain the floor level $z_{\text{floor}}$, we identify frames where the toe joints move below a small speed threshold (quasi-static stance), cluster their toe heights, and take the lowest cluster as $z_{\text{floor}}$. 
All foot $z$-coordinates are then measured relative to this floor level when computing $H_{\text{feet}}$ and foot sliding.

\myparagraph{Foot sliding.}
For the ankles and toes, we accumulate horizontal displacement between consecutive frames, but only when each joint is near the floor, as:
\begin{equation}
    \sum_{t=0}^{T-1}\mathbbm{1}(z_{j,t}<H_j)d_{j,t}(2-2^{z_{j,t}/H_j}),
\end{equation}
\label{eq:foot_sliding}
where $\mathbbm{1}(\cdot)$ is the indicator function, $z_{j,t}$ is the height of joint $j$ at frame $t$, $H_j$ is the joint specific height threshold that defines the near floor region, and $d_{j,t}$ is the horizontal displacement between frame $t$ and $t+1$. The index $j$ runs over the left and right ankles and toes. The accumulated displacement is then averaged over time and across the four joints.

\myparagraph{R-Precision.}
For each text-motion pair, we compute feature vectors for the text and candidate motions, then rank the motions by cosine similarity to the text, following~\cite{guo2022generating, li2024controllable}. $R_{\text{prec}}$ (top-3) is the fraction of pairs for which the ground-truth motion appears in the top-3 ranked motions, averaged over all evaluation pairs.

\myparagraph{FID.}
We compute motion features for all generated and real sequences, fit a Gaussian to each set, and use the Fréchet Inception Distance (FID) between the two Gaussians as the FID score~\cite{heusel2017gans}. 
Lower FID indicates that the distribution of generated motions is closer to that of real motions.

\myparagraph{DIV.}
For each model, we collect motion features from all generated results, sample random feature pairs, and average their Euclidean distances to obtain DIV, following~\cite{guo2022generating}. 
We report a single DIV value per model, where values closer to the DIV of real data indicate more realistic diversity.

\subsection{Interaction Quality}
\myparagraph{Contact metrics.}
For each frame, we assign a binary contact label based on the minimum distance between the hand joints and the object surface: if the closest distance from either hand joint to any object vertex is below $5$\,cm, the frame is marked as in contact. 
Given ground-truth contact labels, $C_\%$ is the fraction of frames predicted in contact, $C_{\text{prec}}$ and $C_{\text{rec}}$ are precision and recall for the contact class, and $C_{F1}$ is their F1 score.

\myparagraph{Penetration metrics.}
We measure mesh interpenetration using signed distance fields (SDFs). At every frame, we map human and object vertices into the corresponding SDF frame and sample their signed distances, with negative values indicating penetration. We average the negative signed distances into four scalars: $P_{\text{hand}}$ and $P_{\text{body}}$ for penetration of hand and full-body vertices into the manipulated object, and $P_{o\rightarrow s}$ and $P_{h\rightarrow s}$ for penetration of the object and the human into the static scene.

\subsection{Ground Truth Difference}

\myparagraph{MPJPE.}
We report MPJPE as the mean Euclidean distance between predicted and ground-truth joint positions~\cite{martinez2017simple}, averaged over all joints and frames in the same world frame.

\myparagraph{Root and object translation error.}
$T_{\text{root}}$ and $T_{\text{obj}}$ are the mean Euclidean distances between predicted and ground-truth root joint positions and object centroids, respectively, averaged over all frames.

\myparagraph{Object orientation error.}
$O_{\text{obj}}$ measures the discrepancy between predicted and ground-truth object rotations.
We compute the mean Frobenius norm of the difference between the two rotation matrices per frame and average over time.

\begin{figure}[t]
    \centering
    \includegraphics[width=\linewidth]{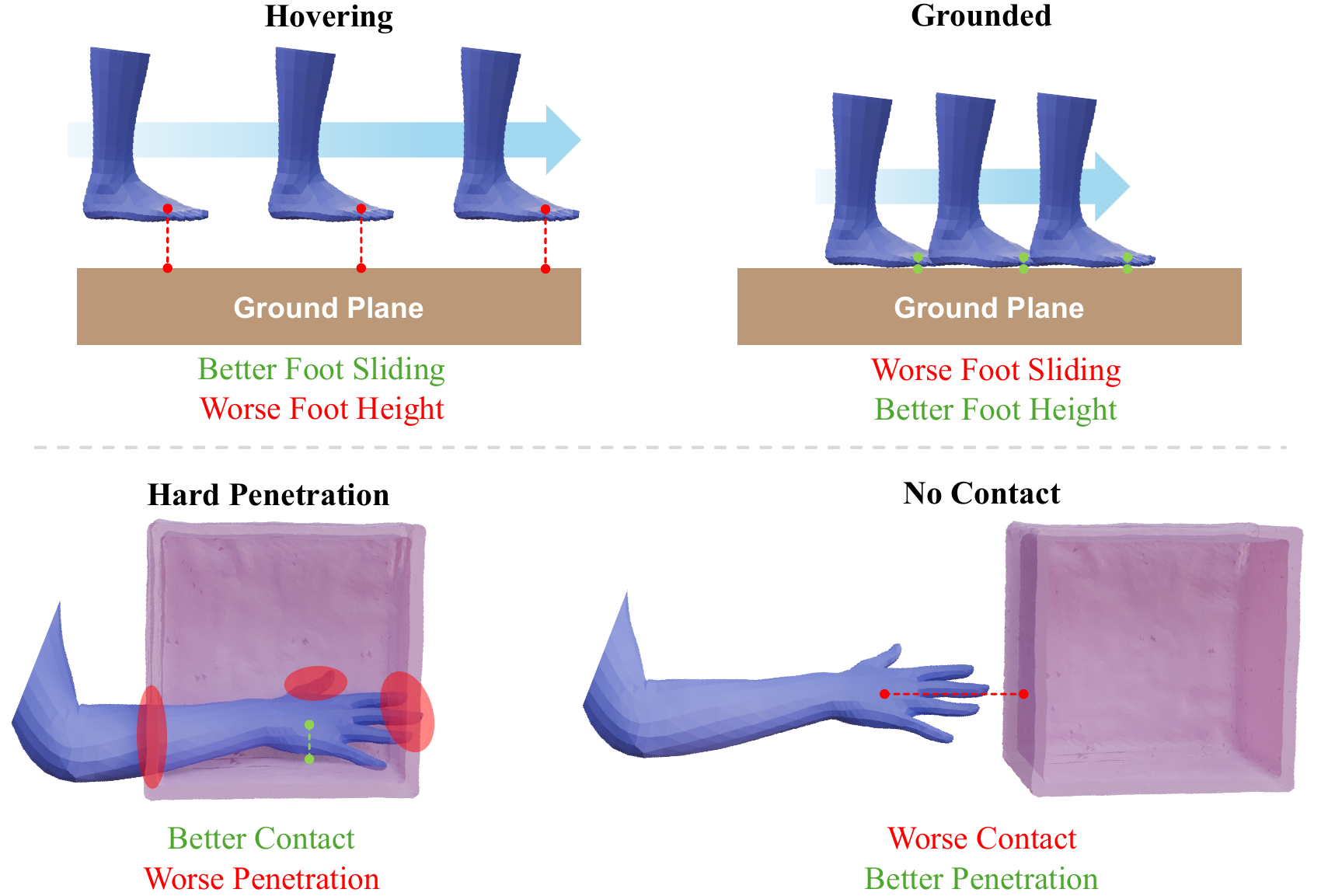}
    \vspace{-1.7em}
   \caption{Visualization of trade-off relationships induced by the metric definitions. The top row illustrates the trade-off between foot related metrics, and the bottom row shows the relationship between contact and penetration metrics.}
    \label{fig:supp_tradeoff}
    \vspace{-1em}
\end{figure}

\subsection{Metrics Trade-off}
\label{supp_sec:metric_trade_off}
\myparagraph{Foot-Ground trade-off.}
Foot sliding ($FS$) assigns larger penalties when $H_{\text{feet}}$ is small, as defined in Eq.~\ref{eq:foot_sliding}. Consequently, as illustrated in the top row of Fig.~\ref{fig:supp_tradeoff}, even large horizontal displacements yield little or no foot sliding when the feet remain high above the ground (hovering). In contrast, when the feet stay very close to the floor, even small displacements are amplified in the $FS$ score. This behavior induces a trade-off between $H_{\text{feet}}$ and $FS$.

\myparagraph{Contact-Penetration trade-off.}
As shown in the bottom row of Fig.~\ref{fig:supp_tradeoff}, the contact-relative score and the penetration score also exhibit a trade-off relationship. Since contact is determined based on the distance between each hand joint and the nearest object vertex, severe penetration can still yield a stable contact score. Conversely, when the hand does not approach the object, the penetration score can be favorable due to the absence of intersecting regions, while the contact score becomes poor. Therefore, metrics with such trade-off behavior should not be interpreted as a single score in isolation but evaluated jointly.

\section{Inference-time Reconstruction Guidance}
\label{sec_c}
To obtain robust and plausible motion during inference, we apply reconstruction guidances following prior work~\cite{li2024controllable}. First, to strengthen hand and foot contacts, we define a guidance term that regularizes the distance between the distal human joints and their nearest object vertices. For the hand joints this term is written as:
\begin{equation}
    \mathcal{F}_{\text{cont}} = \|\textbf{M}_{l} \odot |H_l - V_l|\|_1 + \|\textbf{M}_{r} \odot |H_r - V_r|\|_1,
\end{equation}
where $H_l$ and $H_r$ denote the left and right hand joint positions and $V_l$ and $V_r$ are the corresponding closest object vertices. The binary masks $\mathbf{M}_l$ and $\mathbf{M}_r$ indicate whether a hand joint is in contact with the object and are obtained by thresholding the hand to object distance with a threshold of 5 cm.

Second, to encourage stable foot grounding and a realistic stance, we introduce a guidance term that regularizes the distance between the feet and the floor plane. Given the positions of the left and right feet $F_l$ and $F_r$, this term is defined as:
\begin{equation}
    \mathcal{F}_{\text{feet}} = \|\min(F_l, F_r) - h\|_2,
\end{equation}
where $h = 0.02$ m is a threshold for foot height estimated from the ground-truth motion. The error is computed only along the vertical coordinate.
\begin{figure}[t]
    \centering
    \includegraphics[width=\linewidth]{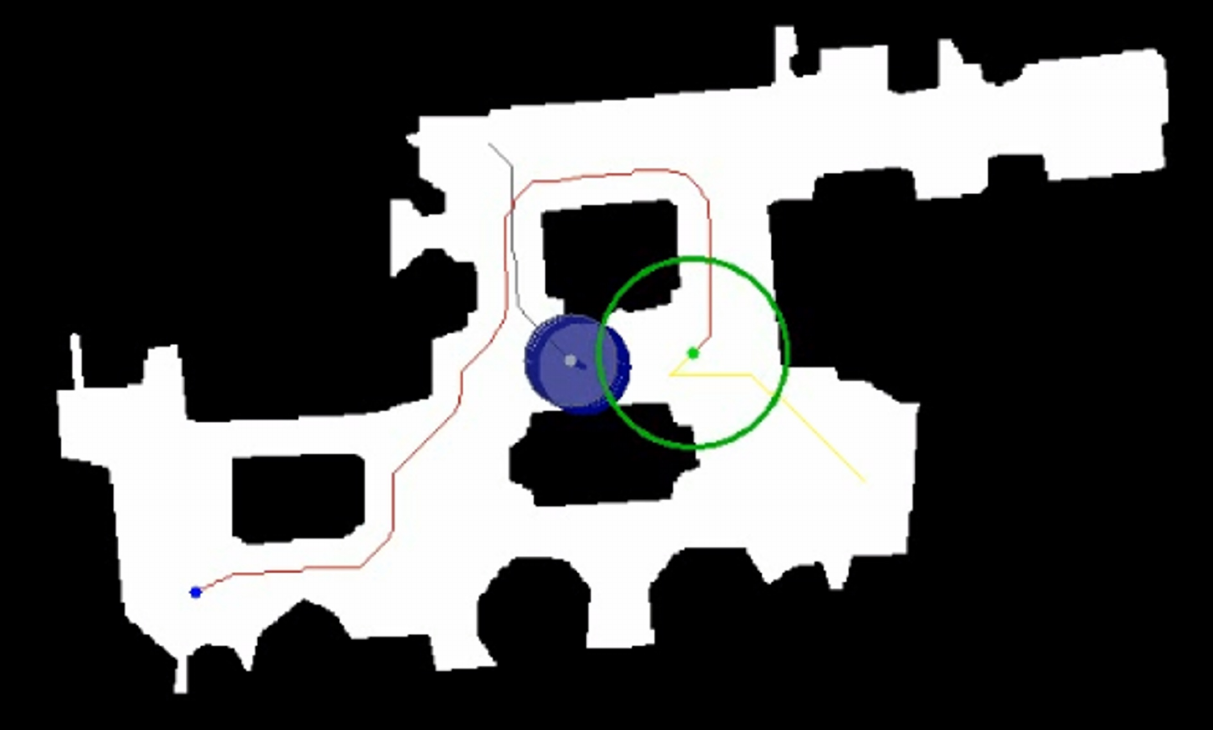}
    \caption{
The green dot and circle denote the agent and $R_{\text{agent}}$, and the red curve shows the re-planned path.
When the obstacle (gray) and its influence region (blue) enter $R_{\text{agent}}$, that area is given high cost in the risk map and dynamic re-planning is triggered.}
    \label{fig:path1}
\end{figure}
\vspace{-2.0em}
\vspace{2em}
\section{DynaPlan}
\label{sec_d}
Inspired by~\cite{li2024controllable}, we project each 3D indoor scene from Replica~\cite{straub2019replica} into a 2D grid layout, marking walkable regions as traversable and expanding non-walkable regions (e.g., walls and furniture) with a distance transform to enforce a human scale safety margin. \textit{DynaPlan} then performs path planning for the agent and obstacle on this grid using A*. After the paths are optimized, the resulting trajectories are lifted back into the original 3D scene using the known metric scale~\cite{li2024controllable, straub2019replica}, and resampled for the agent and obstacles so they provide global plans and motion priors for full-body HOI generation in a consistent 3D coordinate frame. Fig.~\ref{fig:path1} visualizes the overall execution process of \textit{DynaPlan}.

\subsection{Obstacle Trajectory Modeling}
\label{sec:obstacle-trajectory}
Given obstacle start and goal points, the moving obstacle follows a deterministic A*-based trajectory, advancing by one grid cell at each time step $t$. \textit{DynaPlan} uses a pre-trained Social-STGCNN~\cite{mohamed2020social} to forecast future obstacle positions. Given the past $T_{\text{obs}}$ obstacle positions, the model predicts $T_{\text{pred}}$ future positions. These predicted positions do not control the obstacle. Instead, they are used only to construct predictive risk regions that \textit{DynaPlan} uses for planning.

\subsection{Collision Detection}

Given an agent start and goal points, \textit{DynaPlan} first computes an initial path using A*, minimizing the accumulated Euclidean step length ($\sum_{t=1}^{T} \|x_t - x_{t-1}\|_2$). In addition, we assign an influence radius to the agent and obstacle, and declare a collision whenever the obstacle’s radius along its path overlaps with that of the agent. For collision detection we use only the predicted obstacle trajectories. Once a collision is detected, the agent updates a risk map and calls A* again to re-plan its path.

\subsection{Dynamic Re-planning}
\label{sec:dynamic-replanning}
When a potential collision is detected, \textit{DynaPlan} performs dynamic re-planning by updating the risk map based on the current and predicted obstacle positions.
For a grid location $x$, the risk map $R(x)$ is:
\begin{equation}
\begin{aligned}
R(x)
&= 
   \exp\!\left(-\frac{\|x - o_t\|_2^2}{2\sigma^2}\right) \\
&\quad
 + \sum_{k=1}^{T_{\text{pred}}} \ \gamma_{k}
   \exp\!\left(-\frac{\|x - \hat{o}_{t+k}\|_2^2}{2\sigma^2}\right).
\end{aligned}
\end{equation}
We use a Gaussian weight with $\sigma = 1$ to assign smaller values to locations farther from the obstacle, and set $\gamma \in (0,1)$ so that predicted future positions contribute less than the current position.

Given a risk map $R$, the cost of a candidate path $P$ is:
\begin{equation}
\mathcal{C}(P \mid R)
=
\sum_{t=1}^{T}
\left(
\|x_t - x_{t-1}\|_2
+ \lambda_{R} R(x_t)
\right).
\end{equation}
We set $\lambda_{R} = 4.0$ to balance A* cost and risk cost.

To decide between detouring and waiting, \textit{DynaPlan} evaluates candidate waiting time steps ranging from zero (detouring) to $T_{\text{pred}}$.
For each waiting candidate, the agent holds its position for the corresponding duration while the obstacle progresses, and the risk map is updated accordingly.
A waiting time penalty is applied to discourage long waits, by adding to the risk map cost an amount proportional to the number of waiting time steps.

\textit{DynaPlan} selects the candidate with the lowest score, providing a lightweight, prediction-driven re-planning mechanism that enables collision-aware navigation.

\section{Additional Experiments}
\label{sec_e}
\subsection{Ablation Study on Generator Weight}
\begin{table}[t]
\begin{center}
\resizebox{\linewidth}{!}{
\begin{tabular}{@{}lccccccc@{}}
\toprule
& \multicolumn{2}{c}{Condition Matching} 
& \multicolumn{2}{c}{Human Motion Quality} 
& \multicolumn{3}{c}{Interaction Quality} \\ 
\cmidrule(lr){2-3} \cmidrule(lr){4-5} \cmidrule(lr){6-8}
Methods & $T_s$$\downarrow$ & $T_e$$\downarrow$ & FS$\downarrow$ & $R_{prec}$$\uparrow$ & $C_{F1}$$\uparrow$ & $P_{\text{hand}}$$\downarrow$ & $P_{\text{body}}$$\downarrow$ \\ 
\midrule
$\lambda_{\text{G}}$ = 0.01 & 1.69 & 7.60 & 0.45 & 0.70 & \textbf{0.69} & 0.55 & 0.56 \\
$\lambda_{\text{G}}$ = 0.03 & 1.69 & 7.83 & 0.40 & 0.69 & 0.67 & \textbf{0.53} & \textbf{0.54} \\
$\lambda_{\text{G}}$ = 0.05  & 1.98 & 7.87 & 0.42 & 0.69 & 0.66 & \textbf{0.53} & 0.56 \\
\textbf{Ours (DecHOI)} & \textbf{1.59} & \textbf{6.91} & \textbf{0.38} & \textbf{0.72} & 0.67 & \textbf{0.53} & \textbf{0.54}\\
\bottomrule
\end{tabular}
}
\vspace{-0.5em}
\caption{
Ablation comparing fixed and adaptive $\lambda_{\text{G}}$ settings shows that the adaptive strategy yields a better overall balance.
}
\label{tab:ablation_2}
\end{center}
\vspace{-1.5em}
\end{table}

In this section, we investigate the effect of the generator loss weight $\lambda_{\text{G}}$ on the overall objective (Eq. 9) defined in the main paper. Tab.~\ref{tab:ablation_2} compares fixed settings where $\lambda_{\text{G}}$ is set to 0.01, 0.03, or 0.05 with an adaptive strategy that dynamically adjusts $\lambda_{\text{G}}$  between 0.01 and 0.05 according to the performance of the discriminator $\mathcal{D}$. When $\mathcal{D}$ becomes too strong, the adaptive scheme increases $\lambda_{\text{G}}$ so that the generator places more emphasis on fooling the discriminator. When $\mathcal{D}$ influence weakens, the scheme decreases $\lambda_{\text{G}}$ to avoid over-regularization by the adversarial term.

With fixed values, training tends to drift toward an unbalanced regime where either the generator or the discriminator dominates, which often prevents the loss from converging. This lack of convergence not only degrades interaction quality but also harms the overall synthesis performance, leading to weaker condition matching and motion quality. In contrast, the adaptive weighting maintains a better balance in the adversarial training, stabilizes convergence, and yields the best performance across metrics. These results demonstrate that adaptive generator loss weighting effectively mitigates the inherent instability and bias of adversarial training and enables more reliable optimization.

\begin{table*}[t]
\begin{center}
\resizebox{\textwidth}{!}{%
\begin{tabular}{@{}lcccccccccccccccc@{}} 
\toprule
& \multicolumn{2}{c}{Condition Matching} 
& \multicolumn{5}{c}{Human Motion Quality} 
& \multicolumn{5}{c}{Interaction Quality} 
& \multicolumn{4}{c}{GT Difference} \\ 
\cmidrule(lr){2-3} \cmidrule(lr){4-8} \cmidrule(lr){9-13} \cmidrule(lr){14-17}
Methods & $T_s$$\downarrow$ & $T_e$$\downarrow$ & $H_{\text{feet}}$$\downarrow$ & $FS$$\downarrow$ & $R_{prec}$$\uparrow$ & $FID$$\downarrow$ & $DIV$$\rightarrow$ & $C_{\text{prec}}$$\uparrow$ & $C_{\text{rec}}$$\uparrow$ & $C_{F1}$$\uparrow$ & $P_{\text{hand}}$$\downarrow$ & $P_{\text{body}}$$\downarrow$ & MPJPE$\downarrow$ & $T_{\text{root}}$$\downarrow$ & $T_{\text{obj}}$$\downarrow$ & $O_{\text{obj}}$$\downarrow$ \\ 
\midrule
CHOIS~\cite{li2024controllable} & 1.72 & 6.95 & 4.49 & \textbf{0.35} & 0.66 & 0.69 & 8.37 & \textbf{0.80} & 0.64 & \textbf{0.68} & 0.59 & 0.60 & 15.30 & 24.43 & 13.35 & 0.98 \\
HOIFHLI~\cite{wu2025human} & \secbest{1.64} & 8.05 & 4.91 & 0.36 & 0.62 & 1.51 & 8.85 & 0.79 & \textbf{0.65} & 0.67 & 0.60 & 0.58 & 15.93 & \textbf{23.31} & \textbf{10.98} & 1.10 \\
\midrule
\textbf{Ours (DecHOI)} & \textbf{1.59} & \textbf{6.91} & \textbf{4.42} & 0.38 & \textbf{0.72} & \textbf{0.33} & \textbf{8.86} & \textbf{0.80} & 0.64 & 0.67 & \textbf{0.53} & \textbf{0.54} & \textbf{15.27} & 25.47 & 22.96 & \textbf{0.86}\\
\bottomrule
\end{tabular}}
\vspace{-0.6em}
\caption{
Quantitative comparison on the \textit{FullBodyManipulation}~\cite{li2023object} in the privileged waypoint supervised setting, where DecHOI operates without intermediate waypoints and CHOIS~\cite{li2024controllable}, HOIFHLI~\cite{wu2025human} receive sparse intermediate waypoints.
}
\label{tab:privileged_experiment}
\end{center}
\vspace{-2em}
\end{table*}
\subsection{Effect of Privileged Waypoint}
In the main paper, we evaluate DecHOI and prior waypoint-based methods~\cite{li2024controllable, wu2025human} under a fair protocol where all models receive the same inputs. For completeness, we additionally report an experiment on the \textit{FullBodyManipulation}~\cite{li2023object} in which prior methods are evaluated under their original waypoint supervised configuration. In this experiment, DecHOI still operates without any intermediate waypoints, while the prior methods are given waypoints and are thus evaluated in a privileged information setting. The quantitative results of this comparison are summarized in Tab.~\ref{tab:privileged_experiment}. Note that for all baselines we report scores reproduced using the official released implementations, since several metrics are not reported in the CHOIS and HOIFHLI papers (e.g., $DIV$ and $P_{\text{body}}$).

Both CHOIS~\cite{li2024controllable} and HOIFHLI~\cite{wu2025human} benefit noticeably from waypoint supervision. In particular, human motion quality and GT difference metrics improve significantly. The GT difference metrics are highly sensitive to waypoints, and for $T_{\text{obj}}$ we observe improvements of up to a factor of five, since waypoints directly specify object target positions. For this reason, GT difference metrics are not suitable for a direct comparison with DecHOI, which does not receive any waypoint input.

In contrast, DecHOI achieves comparable or better performance than the prior approaches on most metrics, even without privileged conditions. Considering the trade-off structure among metrics discussed in Sec.~\ref{supp_sec:metric_trade_off}, DecHOI shows slightly better overall performance in terms of $H_{\text{feet}}$ and $FS$ and also provides improved text alignment, realism, and diversity, enabled by our decoupled design. Moreover, given the inherent trade-off between contact score and penetration, DecHOI attains superior interaction quality than prior work, despite operating without intermediate waypoints while the baselines use them. These results suggest that DecHOI enables efficient and realistic interaction synthesis under weaker input priors and that the proposed adversarial training makes a clear contribution to improving interaction quality.

\subsection{Oracle Trajectory Ablation}
\begin{table}[t]
\centering
\resizebox{\linewidth}{!}{
\begin{tabular}{@{}lcccccccc@{}}
\toprule
& \multicolumn{3}{c}{Human Motion} 
& \multicolumn{5}{c}{Interaction} \\
\cmidrule(lr){2-4} \cmidrule(lr){5-9}
Methods 

& $R_{\text{prec}}\uparrow$
& $FID\downarrow$
& $DIV$$\rightarrow$
& $C_{\text{prec}}\uparrow$
& $C_{\text{rec}}\uparrow$
& $C_{F1}\uparrow$
& $P_{\text{hand}}\downarrow$
& $P_{\text{body}}\downarrow$ \\
\midrule
DecHOI
& \textbf{0.72} & 0.33 & 8.86 & 0.80 & 0.64 & 0.67 & \textbf{0.53} & \textbf{0.54} \\
DecHOI (oracle)
& 0.66 & \textbf{0.15} & \textbf{8.97} & \textbf{0.82} & \textbf{0.65} & \textbf{0.69} & 0.56 & 0.57 \\
\bottomrule
\end{tabular}
}
\vspace{-0.5em}
\caption{
Oracle trajectory ablation study on the FullBodyManipulation~\cite{li2023object}. DecHOI (oracle) denotes a variant in which the trajectory generator is replaced by ground truth trajectories.
}
\vspace{-1em}
\label{tab:oracle}
\end{table}

We evaluate an oracle setting that bypasses the trajectory generator (TG) and feeds the action generator (AG) with ground-truth object and human trajectories $\{\mathcal{T}_o^{\text{GT}}, \mathcal{T}_h^{\text{GT}}\}$ for all $T$ frames. The AG architecture, checkpoint, and inference-time guidance are kept unchanged, and the conditioning interface follows Sec.~3.2 (see Fig.~2) in the main paper. This design disentangles the contributions of the TG and AG, allowing us to assess the standalone performance of the TG and to estimate the theoretical upper bound of AG performance within the decoupled pipeline. We conduct this experiment on the \textit{FullBodyManipulation}~\cite{li2023object}, and report the results in Tab.~\ref{tab:oracle}.


\medskip
\noindent\textbf{Results and discussion.}
In this oracle configuration, metrics that quantify condition matching or deviation from the conditioning trajectories are less informative, since the AG is driven directly by ground-truth object and human paths. We therefore focus our analysis on human motion quality and interaction quality. The oracle setting generally achieves higher scores on most metrics, yet the gap relative to the standard DecHOI that uses predicted trajectories remains modest. On several metrics ($R_{prec}$, $P_{\text{hand}}$, and $P_{\text{body}}$), the standard DecHOI even slightly outperforms the oracle, and its overall performance is very strong when the typical trade-offs between contact and penetration are taken into account. This small performance difference indicates that the TG generates realistic trajectories that closely follow the distribution of ground-truth paths, and suggests that the AG is not tightly constrained by the TG performance but is capable of producing high quality interaction synthesis independently.
\section{Loss Landscape Visualization}
\label{sec_f}
We visualize the loss landscape around our converged model by evaluating the training objective on a two-dimensional subspace of the parameter space, following~\cite{li2018visualizing}. We fix a batch $\mathcal{B}$ drawn from the same distribution as the training data, so that variations in the loss reflect only changes in the model parameters.
Let $w_0$ be the vector obtained by flattening and concatenating all trainable tensors of the final trained model.

To explore the neighborhood of $w_0$, we construct two random perturbation directions. For each trainable tensor, we sample Gaussian noise with the same shape, normalize it to obtain a comparable scale across layers, and concatenate these tensors to form the first direction $d_x$. We repeat this procedure to obtain a second direction $d_y'$ and orthogonalize it with respect to $d_x$ using Gram-Schmidt to obtain the final direction $d_y$. This keeps the perturbations balanced across layers and avoids a single layer dominating the change.

We then form a grid of coefficients $\alpha, \beta \in [-r, r]$ and, for each pair $(\alpha, \beta)$, construct a perturbed parameter vector $w(\alpha, \beta) = w_0 + \alpha d_x + \beta d_y$, reshape it back to the original tensor shapes, and evaluate the total loss on $\mathcal{B}$. The guidance term is omitted since it is used only at inference time and is not part of the training objective. Plotting the resulting loss values over the $(\alpha, \beta)$ grid as contour and surface plots provides a qualitative view of the local loss landscape around $w_0$ and indicates whether the optimization problem near the solution is relatively simple with a smooth basin or more complex with sharper curvature and narrower valleys.
\section{OMOMO Implementation Details}
\label{sec_g}
We implement Lin-OMOMO and Pred-OMOMO within the original OMOMO framework~\cite{li2023object}, following the variant definitions introduced in CHOIS~\cite{li2024controllable}. To match the conditioning used in our experiments, our OMOMO variants are provided only with the object start and goal states.


\subsection{Lin-OMOMO}
Lin-OMOMO uses the original OMOMO generator and training setup. The object trajectory is defined by linearly interpolating the object centroid between its start and goal positions and is used as the object translation input in the OMOMO conditioning stream. All centroid-related values are updated accordingly. Since Lin-OMOMO only interpolates the object centroids and keeps the orientation identical to the ground-truth, we omit $O_{\text{obj}}$ for Lin-OMOMO from our reported results.

\subsection{Pred-OMOMO}
In Pred-OMOMO, the original OMOMO generator and training configuration are kept intact, while the object motion is obtained from CHOIS. Given the object's start and goal conditions as input, CHOIS predicts the full trajectory of object centroids and rotations. We use this predicted object motion as the input in the OMOMO conditioning stream and update all related pose values to stay in sync with these predictions. Because both the trajectory of the object centroid and its rotations are supplied directly by CHOIS, the object-level metrics $T_{\text{obj}}$ and $O_{\text{obj}}$ for Pred-OMOMO are identical to those of CHOIS in our experiments.

\section{User Study Details}
\label{sec_h}
\begin{figure}[t]
    \centering
    \includegraphics[width=\linewidth]{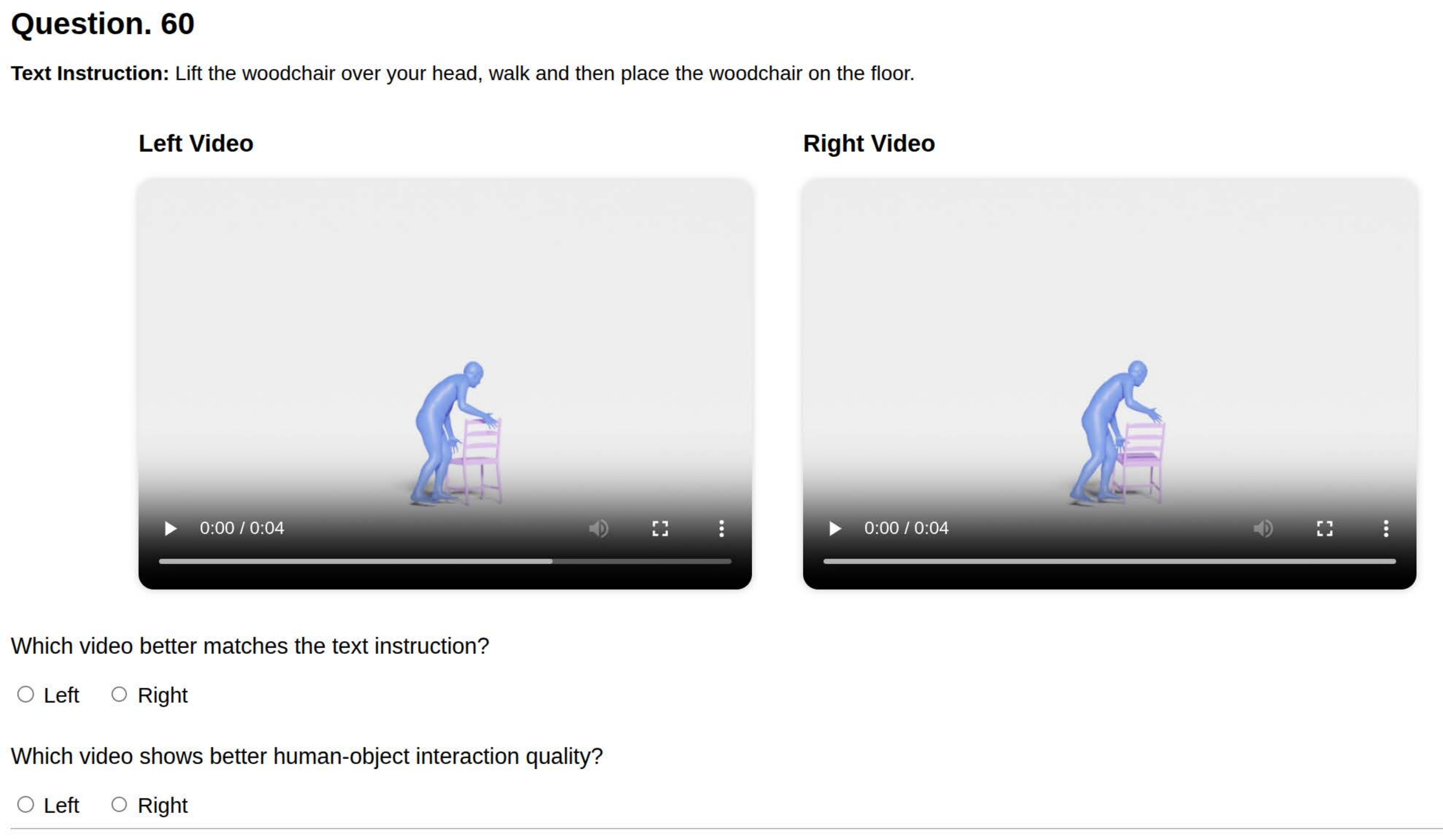}
    \vspace{-1.6em}
   \caption{Example 2AFC interface in which participants read a text instruction and compare two anonymized clips to judge text alignment and interaction quality.}
    \label{fig:user_supp}
    \vspace{-1em}
\end{figure}
We conducted a two-alternative forced-choice (2AFC) perceptual study to evaluate two criteria of synthesized human-object interactions: Text Alignment (how faithfully a clip follows the natural-language instruction) and Interaction Quality (the perceived realism and plausibility of contact, including stable hand and foot contacts, few penetrations, and minimal temporal jitter).

For each comparison round (DecHOI-CHOIS~\cite{li2024controllable} and DecHOI-HOIFHLI~\cite{wu2025human}), we randomly sampled 30 text-scene pairs from the full test set without scene duplication, yielding 60 clips in total. To ensure fairness, we shuffled both the order of videos and the left-right placement of each method in every trial, and fixed all video durations to 4 seconds. As shown in Fig.~\ref{fig:user_supp}, the interface presents the Text Instruction at the top and two anonymized videos. Participants could replay videos and answer two questions. They selected which clip better matched the instruction (Text Alignment) and which clip exhibited better human-object interaction quality (Interaction Quality). To avoid presentation bias, no method identifiers or cues were provided.

Using the results in Fig.~8 of the main paper, participants preferred DecHOI in 71.5\% of trials versus CHOIS and 67.5\% versus HOIFHLI for Text Alignment, and in 69.0\% of trials versus CHOIS and 63.5\% versus HOIFHLI for Interaction Quality. These trends are consistent with our quantitative metrics and support the claim that decoupling trajectory and action generation improves both semantic faithfulness and perceptual interaction quality.

\section{Additional Qualitative Results}
\label{sec_i}
We provide additional qualitative results that complement the examples in the main paper. 

\begin{figure*}[t]
    \centering
    \includegraphics[width=1.0\textwidth]{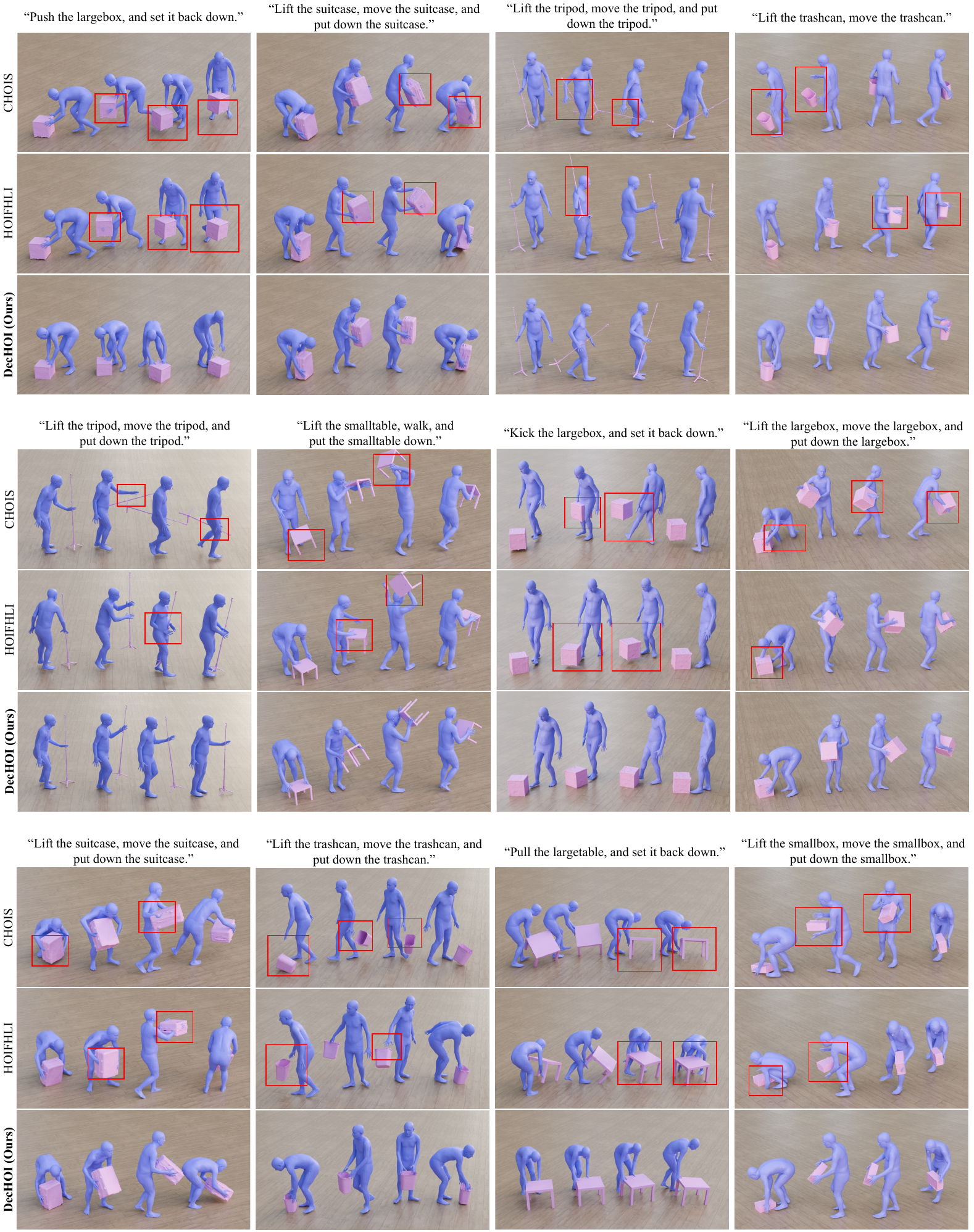}
    \vspace{-1.9em}
    \caption{
    Additional qualitative comparison of DecHOI with CHOIS~\cite{li2024controllable} and HOIFHLI~\cite{wu2025human} on the \textit{FullBodyManipulation}~\cite{li2023object}. Also DecHOI produces stable contacts, smooth motion, and accurate object trajectories, while prior methods show drift, penetration, or inconsistent coordination between human and object motions. 
    }
    \label{fig:supp_quality}
    \vspace{-1.7em}
\end{figure*}

Fig.~\ref{fig:supp_quality} extends Fig.~4 in the main paper by including additional scenes from \textit{FullBodyManipulation}~\cite{li2023object}, comparing DecHOI against CHOIS~\cite{li2024controllable} and HOIFHLI~\cite{wu2025human}. Across diverse instructions (e.g., lift-move-place, push and pull), DecHOI maintains stable hand-object contacts and grounded object trajectories, exhibiting less penetration and hovering. In contrast, the baselines show drift, temporal desynchronization between human and object, or incomplete contacts, especially during lift-place transitions and when objects change orientation.

\begin{figure*}[t]
    \centering
    \includegraphics[width=1.0\textwidth]{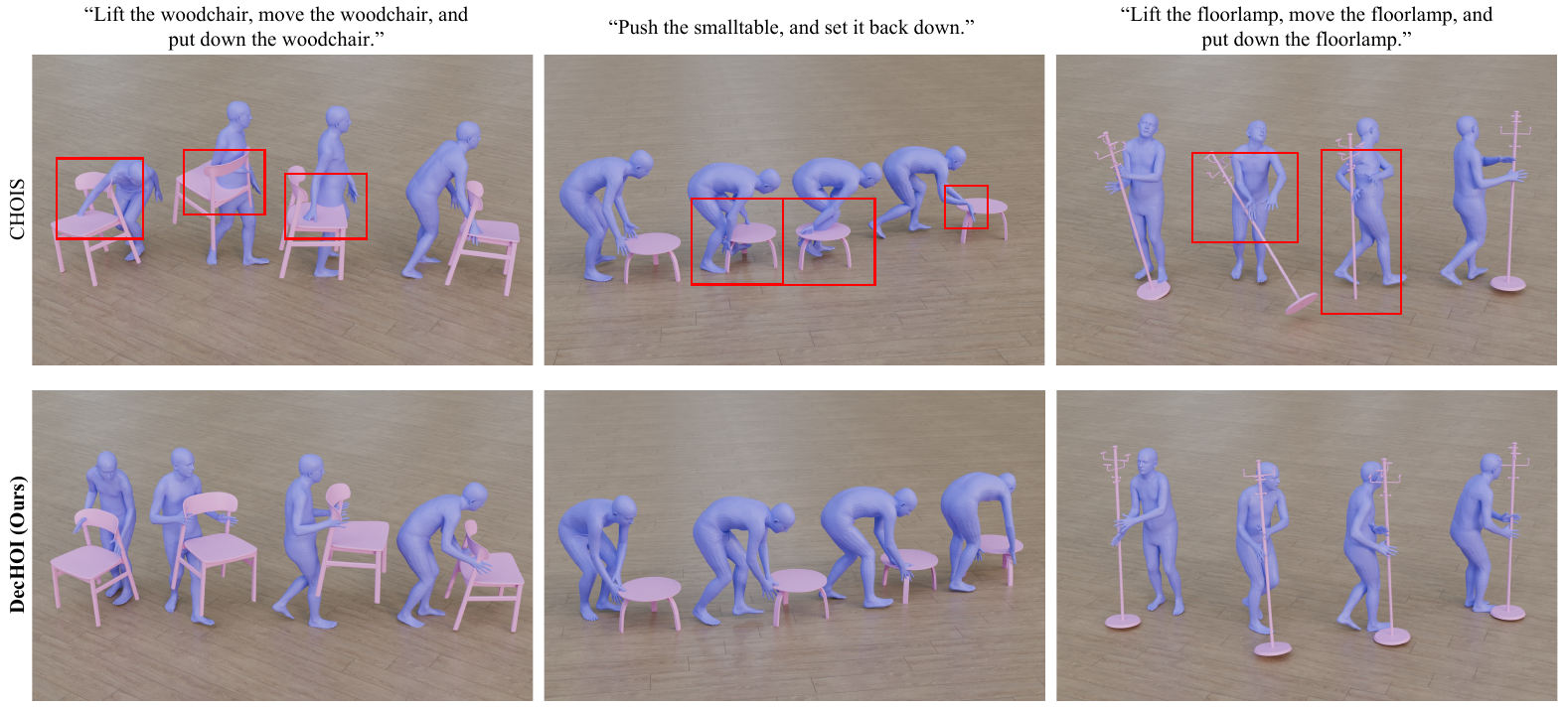}
    \vspace{-1.9em}
    \caption{
    Additional qualitative comparison of DecHOI and CHOIS~\cite{li2024controllable} on the \textit{3D-FUTURE}~\cite{fu20213d}, demonstrating generalization to unseen object categories such as woodchair, smalltable, and floorlamp.
    }
    \label{fig:supp_unseen_quality}
    \vspace{-1.7em}
\end{figure*}

Fig.~\ref{fig:supp_unseen_quality} provides additional qualitative results that complement Fig.5 in the main paper, showcasing \textit{3D-FUTURE}~\cite{fu20213d} objects that are unseen during training. DecHOI maintains precise hand-object contact locations, more coherent placement motion, and consistent text to motion alignment, whereas CHOIS often produces mesh intersections or unstable object poses. The qualitative patterns align with the quantitative improvements reported for condition matching, motion stability, and contact reliability on unseen shapes.

\begin{figure*}[t]
    \centering
    \includegraphics[width=1.0\textwidth]{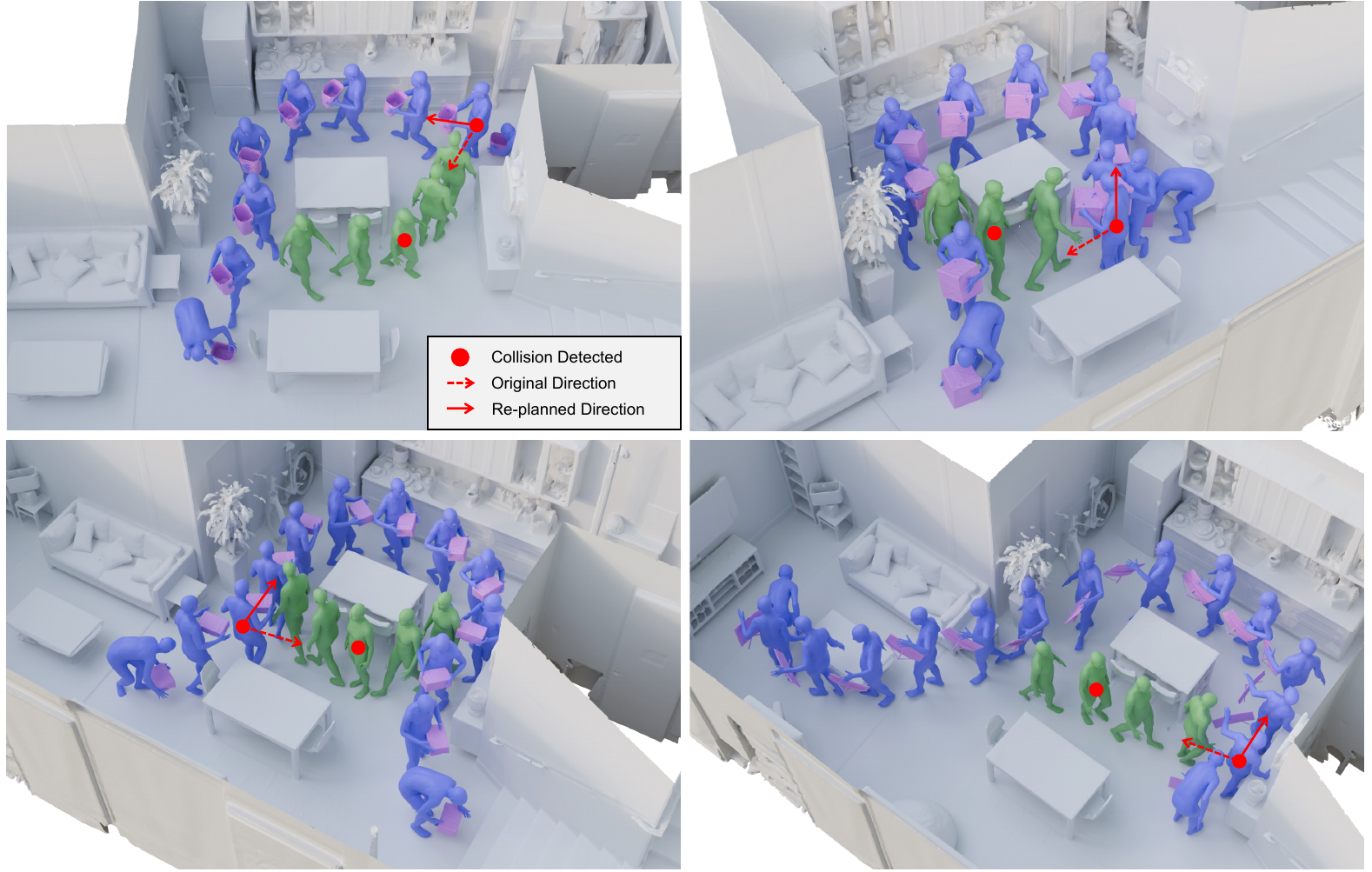}
    \vspace{-1.9em}
    \caption{
   Visualization of DecHOI in long-sequence dynamic environments. The human agent (blue) adaptively re-plans its path when encountering a moving obstacle (green), often waiting briefly before detouring, and thereby maintaining goal-directed and collision-free motion.
    }
    \label{fig:supp_dynaplan_quality}
    \vspace{-1.7em}
\end{figure*}

Fig.~\ref{fig:supp_dynaplan_quality} complements Fig.~7 in the main paper with additional long-horizon dynamic scenes from \textit{DynaPlan}. We visualize collision events together with the corresponding re-planned direction. The dashed arrows indicate the original direction of motion, which would be the optimal path in free space, yet the agent re-routes around obstacles instead of strictly following this direction. These behaviors demonstrate that obstacle detection and dynamic planning are applied effectively and suggest that the framework can scale to more complex scenarios.
\section{Limitations \& Future Work}
\label{sec_j}
Our framework focuses on synthesizing human-object interactions that involve manipulating rigid objects. In real environments, however, many objects are deformable or articulated. Handling such cases requires contact regions that adapt over time to object parts whose positions change, and in some scenarios it is necessary to model underlying physical effects such as friction, inertia, and gravity. Addressing these challenges would require more sophisticated representations of articulated structure and dynamics. Incorporating recent articulation estimation networks as object priors is a promising direction, and systematically extending our approach to deformable and articulated objects constitutes an important avenue for future work.

In addition, the proposed \textit{DynaPlan} module currently supports dynamic planning and interaction synthesis for a single manipulated object and a single moving obstacle. Real-world scenarios often involve more complex environments, such as crowds of agents and multiple objects being manipulated simultaneously. Scaling to such settings may require path planners that go beyond classical A* and can reason about many interacting agents. Moreover, supporting multiple manipulated objects would likely depend on constructing richer datasets, for example by combining motion capture with procedural composition of multi object interactions. Exploring these extensions would be highly valuable for broadening the applicability of our framework.
\clearpage
\clearpage 

{\small
\bibliography{supp_main,main}
}